\documentclass[times,twocolumn,review]{elsarticle}

\usepackage{medima}
\usepackage{framed,multirow}

\usepackage{amssymb}
\usepackage{latexsym}
\usepackage{amsmath}
\usepackage{graphicx}

\usepackage{url}
\usepackage{xcolor}
\usepackage{booktabs}
\usepackage{hyperref}
\usepackage{subcaption}
\usepackage{amsmath}

\definecolor{newcolor}{rgb}{.8,.349,.1}

\journal{}

\begin{document}

\verso{Yazıcı \textit{et~al.}}

\begin{frontmatter}

\title{VEELA: A Clinically-Constrained Benchmark for Liver Vessel Segmentation in Computed Tomography Angiography}%

\author[1]{Ziya Ata \snm{Yazıcı}\corref{cor2}}
\author[2]{N. Sinem \snm{Gezer}}
\author[3]{İlkay \snm{Öksüz}}
\author[4]{İlker Özgür \snm{Koska}}
\author[4]{Tuğçe \snm{Toprak}}
\author[4]{Pervin \snm{Bulucu}}
\author[4]{Ufuk \snm{Beşenk}}
\author[4]{A. Emre \snm{Kavur}}
\author[5]{Pierre-Henri \snm{Conze}}
\author[3,6]{Hazım Kemal \snm{Ekenel}}
\author[2]{Oğuz \snm{Dicle}}
\author[8]{Mustafa Ege \snm{Şeker}}
\author[9]{Mustafa Said \snm{Kartal}}
\author[10]{Ariorad \snm{Moniri}}
\author[8]{Orhan \snm{Özkan}}
\author[11]{Osman Faruk \snm{Bayram}}
\author[12]{Hakan \snm{Polat}}
\author[13]{Musa \snm{Balcı}}
\author[13]{Ece Tuğba \snm{Cebeci}}
\author[13]{Baran \snm{Cılga}}
\author[13]{Kardelen \snm{Peçenek}}
\author[7]{M. Alper \snm{Selver}\corref{cor1}}

\cortext[cor1]{Corresponding author:
e-mail: alper.selver@deu.edu.tr}
\cortext[cor2]{Co-corresponding author:
e-mail: ziya.ata.yazici@huawei.com}

\address[1]{Huawei Türkiye R\&D Center, Istanbul, 34768, Türkiye}
\address[2]{Department of Radiology, Dokuz Eylul University, Izmir, 35100, Türkiye}
\address[3]{Department of Computer Engineering, Istanbul Technical University, Istanbul, 34469, Türkiye}
\address[4]{Institute of Natural and Applied Sciences, Dokuz Eylul University, Izmir, 35100, Türkiye}
\address[5]{IMT Atlantique, LaTIM UMR 1101, Inserm, Brest, 29200, France}
\address[6]{Division of Engineering, NYU Abu Dhabi, 129188, UAE}
\address[7]{Department of Electrical and Electronics Engineering, Dokuz Eylul University, Izmir, 35100, Türkiye}
\address[8]{Department of Radiology, University of
Wisconsin-Madison, Madison, WI, 53792, USA}
\address[9]{School of Medicine, Sivas Cumhuriyet
University, Sivas, 58140, Türkiye}
\address[10]{School of Medicine, Acibadem Mehmet Ali Aydinlar
University, Istanbul, 34638, Türkiye}
\address[11]{Department of Artificial Intelligence Engineering, Bahçeşehir University, Istanbul, 34349, Türkiye}
\address[12]{Faculty of Pharmacy, Sivas Cumhuriyet University, Sivas, 58140, Türkiye}
\address[13]{Viseur AI, Ankara, 06510, Türkiye}

\received{}
\finalform{}
\accepted{}
\availableonline{}
\communicated{}

\begin{abstract}

Accurate segmentation of hepatic and portal vessels in contrast-enhanced computed tomography angiography (CTA) remains challenging due to complex vascular topology, peripheral visibility limitations, and acquisition-induced ambiguities. While existing public datasets offer valuable benchmarks, few include clinically realistic annotation constraints. We introduce VEELA (Vessel Extraction and Extrication for Liver Analysis), a rigorously curated liver vessel dataset derived from 40 CTA scans inherited from the CHAOS grand-challenge cohort. All vessels were manually delineated slice-by-slice under multi-expert consensus, using a strict visibility-driven annotation policy and avoiding anatomically inferred interpolation. This design explicitly captures anatomical variability and imaging-related uncertainty. As a continuation of the CHAOS challenge, VEELA enables reproducible cross-benchmark evaluation while extending the scope to fine-grained hepatic and portal vessel segmentation. We further establish a standardized benchmarking framework and analyze complementary evaluation metrics, including topology-aware (clDice), overlap-based (IoU), boundary-sensitive (NSD), and geometry-aware (area, length) measures. Our results demonstrate that different metrics capture distinct aspects of vascular integrity, underscoring the necessity of multi-perspective evaluation for clinically meaningful vessel segmentation. VEELA is publicly released to facilitate reproducible research and support the development of robust vascular segmentation methods. Researchers can access the evaluation metrics, dataset, and submission platform at \url{https://www.synapse.org/Synapse:syn65471967}.

\end{abstract}


\end{frontmatter}

\section{Introduction}
\label{introduction}
Accurate analysis of liver vasculature in three dimensions (3D) is essential for a variety of medical procedures, including preoperative planning, treatment of hepatic diseases, and computer-aided diagnosis. Moreover, it is also used for medical education, which significantly benefits from advanced models and realistic simulations. 

The clinical importance of the liver vessel extraction can be explained through some particular applications:
\begin{enumerate} 
\item Living donated liver transplantation is a surgical operation where part of the donor’s liver is transplanted to another patient. Pre-surgical planning for transplantation requires precise knowledge of the liver vascular morphology \citep{selver2008patient}. Furthermore, the donor’s suitability for the operation can be assessed by performing a volumetric approximation of the liver with vasculature analysis, thereby providing insight into postoperative liver function \citep{selle2002analysis}. 
\item Localization of the liver lesions is based on the lesion’s position relative to the surrounding hepatic vessels \citep{fasel1998segmental}. Hepatocellular carcinoma is the most common liver cancer and is the primary source of cancer-related deaths worldwide \citep{balogh2016hepatocellular}. The most common therapy for hepatocellular carcinoma is radio-frequency treatment. Blood vessels around the tumor can disturb the thermal process by acting as coolers. Therefore, knowledge of the vessels adjacent to the tumor is crucial for determining the most suitable path for inserting an interstitial applicator \citep{esneault2009liver}. 
\item In the late stages of liver cancer, liver resection remains the only curative treatment, requiring surgeons to remove the diseased tissue. Such an operation necessitates the patients to meet a list of preconditions. Patients’ suitability for the procedure depends on the tumor position and size, as well as the expected postoperative liver function, which is another critical factor to consider \citep{reitinger2006liver}.
\end{enumerate}
 
A complete analysis of the liver vasculature implies the analysis of three sub-systems \citep{abdel2010liver}:
\begin{enumerate}
\item Hepatic vessels that transport de-oxygenated blood from the liver to the heart,
\item Portal vessels that transport nutrient-rich blood from the intestines to the liver,
\item The hepatic artery that carries oxygenated blood from the heart to the liver.
\end{enumerate}
Occlusion of one of these vessels can obstruct the drainage or supply for some liver parts \citep{lehmann2008portal}. 

Computed tomography angiography (CTA) is the most common imaging technique for imaging the liver vasculature. An intravascular contrast agent is injected into the patient before the scan, causing vessels to appear brighter than the liver parenchyma. From an image-processing perspective, this process enhances intensity in blood-filled regions and makes segmentation easier, especially for the main branches.

In routine clinical workflow, vessel segmentation is primarily performed by radiologists, who manually annotate all vessels on abdominal multi-slice CT slices. This process is tedious and time-consuming, and prone to high intra- and inter-expert variability. Therefore, automatic liver vessel segmentation is an area of interest. The hepatic vasculature exhibits a complex multi-scale geometry with substantial inter-patient variability. Additionally, the presence of lesions (e.g., tumors, metastases) and prior liver surgery can alter the normal vascular structure \citep{conversano2011hepatic}.

In this respect, the contributions of this paper are:
\begin{enumerate}

\item We introduce VEELA, a rigorously curated and clinically enriched liver-vessel segmentation dataset derived from 40 contrast-enhanced CT angiography (CTA) scans from the CHAOS cohort \cite{kavur2021chaos}. Rather than focusing on dataset scale, VEELA emphasizes annotation fidelity and clinical realism. All vascular structures were manually delineated slice-by-slice under multi-expert consensus, explicitly encoding peripheral visibility constraints, anatomical variations, and acquisition-induced ambiguities. This visibility-driven annotation strategy produces a highly detailed and methodologically transparent vascular ground truth that reflects the practical challenges encountered in real-world clinical imaging.

\item VEELA extends the CHAOS benchmarking system from organ segmentation to the considerably more demanding task of vascular structure extraction. By retaining the same clinical imaging cohort while introducing detailed hepatic and portal vessel annotations, VEELA enables controlled investigation of how segmentation approaches that perform well on organ-level structures generalize to thin, branching vascular networks. This continuity allows methodological robustness to be studied across increasing anatomical complexity, under consistent imaging conditions.

\item We establish a standardized benchmarking framework through the VEELA challenge, which was hosted at IEEE MLSP 2025 as a one-time event with a continuous evaluation platform on Synapse. Our analysis in this work focuses on complementary evaluation metrics for vessel segmentation. By jointly examining topology-aware (clDice), overlap-based (IoU), boundary-sensitive (NSD), and geometry-aware (area, length) measures, we demonstrate how different metrics capture distinct aspects of vascular integrity and structural correctness. This multi-metric analysis highlights limitations of single-score reporting and promotes a more comprehensive and clinically meaningful evaluation paradigm for vessel segmentation.

\end{enumerate}


\section{Related Work}
\label{relatedwork}

This section reviews the existing literature on liver vessel segmentation, organized into four thematic areas: Liver Vessel Segmentation Datasets, Classical and Hybrid Approaches, Deep Learning Methods, and Evaluation Metrics for Liver Vessel Segmentation.

\subsection{Liver Vessel Segmentation Datasets}
\label{subsec:rw_datasets}

Publicly available datasets for liver vessel segmentation remain scarce compared to those for liver parenchyma or tumor delineation. The 3D-IRCADb repository \citep{soler20103d} provides 20 contrast-enhanced CT volumes with annotations of hepatic vessels, liver, and tumors, though its limited size primarily limits its use to validation rather than training. The Medical Segmentation Decathlon (MSD) Task~08 \citep{antonelli2022medical} offers a substantially larger collection of 443 portal-venous-phase CT scans with semi-automatically generated hepatic vessel and tumor labels. More recently, the Liver Vessel Segmentation (LiVS) dataset \citep{gao2023laplacian} introduced 532 contrast-enhanced CT volumes with dedicated liver-vessel annotations (approximately 30 annotated slices per volume), along with a Laplacian salience-gated feature pyramid network that addresses the extreme class imbalance between vessel and background voxels.

Recognizing the lack of multi-class hepatic vessel annotations, two concurrent 2024--2025 efforts have introduced datasets with separate labels for hepatic veins (HV) and portal veins (PV). \citet{xu2024dualstream} re-annotated 273 volumes from MSD Task~08 and introduced the Dual-stream Hepatic Portal Vein Segmentation Network, constructing two sub-datasets stratified by slice thickness (threshold of 2~mm). The AIMS-HPV-385 dataset \citep{cavicchioli2025} introduced a newly collected set of 385 CT volumes with multi-class annotations and accompanied the release with a dual-stage dual-class segmentation framework and a separate vascular connectivity correction (VCC) postprocessing step that explicitly addresses disconnected vessel branches. Together, these datasets highlight a progressive shift from binary vessel masks toward class-discriminative annotations that support Couinaud-based surgical planning.

These datasets differ substantially in annotation methodology and resulting ground-truth fidelity. The MSD Task~08 labels were generated semi-automatically using level-set initialization followed by manual correction \citep{antonelli2022medical}, a process noted to contain errors, particularly for small peripheral branches \citep{xu2024dualstream}. The 3D-IRCADb annotations, while fully manual, were reported to contain approximately 16\% unlabeled vessels \citep{huang2018}. The LiVS dataset \citep{gao2023laplacian} provides manual annotations for only a subset of slices per volume (approximately 30 slices), leaving intermediate regions unannotated. Our dataset, VEELA, advances this trajectory by providing visibility-driven, multi-expert consensus annotations of both hepatic and portal veins on the same 40 portal-venous-phase CT volumes used in the CHAOS challenge \citep{kavur2021chaos}, enabling cross-benchmark compatibility. 

In contrast to prior efforts, VEELA provides slice-by-slice manual annotations for all 40 volumes without any interpolation, reviewed under a multi-expert consensus protocol, and governed by a strict visibility-driven policy that encodes epistemic uncertainty rather than assuming anatomical continuity (see Section~\ref{datasetcoll} for details). As Table~\ref{tab:rw_datasets} illustrates, this constitutes a notably conservative annotation protocol among existing liver vessel segmentation datasets. Preliminary results of the VEELA challenge were presented in \citep{toprak2025veela}; this paper substantially extends that work by detailing the imaging challenges and annotation strategies, introducing a comprehensive benchmarking framework, and conducting a multi-metric evaluation.

\begin{table*}[!t]
\centering
\caption{Liver vessel segmentation datasets. HV = hepatic vein, PV = portal vein, IVC = inferior vena cava. $\dagger$~Re-annotation of MSD Task~08. $\ddagger$~Volumes sourced from CHAOS challenge \citep{kavur2021chaos}. $\S$~Available upon request.}
\label{tab:rw_datasets}
\resizebox{\textwidth}{!}{%
\begin{tabular}{@{}llclllllp{3.2cm}@{}}
\toprule
\textbf{Dataset} & \textbf{Year} & \textbf{Vols} & \textbf{Modality} & \textbf{Phase} & \textbf{Vessel Cls} & \textbf{Other} & \textbf{Pub} & \textbf{Annotation Method} \\
\midrule
3D-IRCADb \citep{soler20103d}          & 2010 & 20  & CE-CT & Portal ven. & HV, PV        & Liver, tumour & \checkmark & Manual \\
MSD Task 08 \citep{antonelli2022medical}    & 2022 & 443 & CE-CT & Portal ven. & Binary        & Tumour        & \checkmark & Semi-auto (level-set) \\
LiVS \citep{gao2023laplacian}                 & 2023 & 532 & CE-CT & Not specified          & Binary        & ---           & \checkmark & Manual \\
Xu et al.$^\dagger$ \citep{xu2024dualstream} & 2024 & 273 & CE-CT & Portal ven. & HV, PV & ---           & \checkmark & Manual re-annotation \\
AIMS-HPV-385 \citep{cavicchioli2025} & 2025 & 385 & CE-CT & Portal ven. & HV, PV        & ---           & \texttimes$^\S$ & Manual \\
Li et al. \citep{li2024automated}       & 2024 & 515 & CE-CT & Portal ven. & HV, PV, IVC   & Liver         & \texttimes & Manual (double-ref.) \\
\textbf{VEELA (ours)}$^\ddagger$     & 2025 & 40  & CE-CT & Portal ven. & HV, PV        & Liver$^\ddagger$ & \checkmark & Manual (Visibility-driven, multi-expert) \\
\bottomrule
\end{tabular}%
}
\end{table*}

\subsection{Classical and Hybrid Approaches}
\label{subsec:rw_classical}

Early liver vessel segmentation methods relied on handcrafted features combined with energy minimization or graph-based optimization. \citet{esneault2009liver} used geometrical moments to detect and model tubular structures and coupled them with graph cuts for final vessel delineation. \citet{zeng2017} proposed a two-stage pipeline combining oriented flux symmetry and oriented flux antisymmetry (OFA) with graph cuts for vessel extraction. \citet{sangsefidi2018} introduced a balanced graph-cut formulation that improved segmentation of thin vessels by re-weighting terminal and source capacities. Building on an existing benchmark of six vesselness enhancement paradigms, Frangi \citep{frangi1998multiscale}, Sato \citep{sato1998}, Jerman \citep{jerman2016enhancement}, Zhang, Meijering, and RORPO, \citet{garret2024deep} fused five of them (excluding Jerman) with a convolutional network, demonstrating that classical multi-scale vessel enhancement still provides complementary information to learned features. \citet{alirr2023} systematically evaluated how different preprocessing pipelines (coherence-enhancing diffusion filtering, vesselness filtering) affect subsequent deep learning performance for liver vessel segmentation, providing empirical guidance on optimal enhancement strategies.

These classical approaches share a common limitation: their reliance on handcrafted priors (tubularity, intensity profiles, local curvature) makes them sensitive to the imaging conditions; contrast timing, noise, and partial volume effects, which are particularly variable in liver CTA. Such challenges are explicitly represented in the VEELA dataset, which includes cases with windmill artifacts, isodense lumen regions, respiratory motion, and variable contrast enhancement (see Section~\ref{datasetcoll} for detailed examples).

\subsection{Deep Learning Methods}
\label{subsec:rw_deep}

The transition to fully data-driven approaches for liver vessel segmentation has been driven by two core challenges: extreme class imbalance between vessel and background voxels, and the multi-scale nature of hepatic vasculature spanning from thick central trunks to thin peripheral branches. To address class imbalance, \citet{huang2018} introduced a variant Dice loss within a 3D U-Net framework, establishing volumetric processing as a viable alternative to earlier slice-wise methods such as the multi-pathway 2D CNNs of \citet{kitrungrotsakul2019}, which operated across three orthogonal planes. To capture multi-scale vessel geometry, \citet{hao2022} proposed HPM-Net, a hierarchical progressive network that acquires receptive fields of different sizes to capture semantic information across spatial scales, while \citet{yan2021} introduced LVSNet, with attention-guided multi-scale feature fusion, that adaptively selects useful low-level features, thereby improving thin-branch detection. Despite these advances, purely convolutional architectures remain constrained by their local receptive fields, limiting their ability to model the long-range spatial dependencies inherent to branching vascular trees.

Transformer-based and hybrid architectures have been explored specifically to overcome this receptive field limitation by modeling long-range spatial dependencies. IBIMHAV-Net \citep{wu2023hepatic} combined convolutional blocks with 3D Swin Transformer blocks, capturing global context that CNNs with limited receptive fields struggle to represent. TransRAUNet \citep{transraunet2025} augmented a TransUNet architecture with a Reverse Attention Module and multi-Hounsfield-unit windowing augmentation specifically designed for hepatic vessel contrast enhancement. GLIMS \citep{yazici2024glims} offers a lightweight hybrid design combining dilated convolutions with a Swin Transformer bottleneck; its architectural details are provided in Section~\ref{propsedmetbaseline}, where it serves as one of the baseline models.

Meanwhile, purely convolutional designs have continued to advance by addressing inter-slice and inter-scale contextual modeling without transformer components. HI-Net \citep{liu2024hi} uses hierarchical multi-scale feature fusion with parallel dilated convolutions and inter-scale dense connections, while SCAN \citep{zhou2024scan} captures inter-slice context via a slice-level attention module and graph association module within a sequence-based context-aware architecture. These approaches demonstrate that carefully designed convolutional mechanisms can partially compensate for the absence of explicit long-range modeling.

Multi-task and vessel-specific architectures represent a further specialization. VSNet \citep{xu2025vsnet} jointly learns voxel-wise segmentation, centerline regression, and edge segmentation through a vessel-growing decoder to enforce topological consistency; it is also employed as a baseline model in this study (see Section~\ref{propsedmetbaseline} for architectural details). \citet{li2024automated} developed a coarse-to-fine algorithm using a private dataset of 515 portal venous phase CT images (413 training), segmenting hepatic veins, portal veins, and IVC separately, achieving DSC values of 0.86, 0.89, and 0.94, respectively.

Three converging trends have emerged in recent literature, reflecting the field's increasing maturity. First, multi-class hepatic--portal segmentation has become a primary focus: the Dual-stream network of \citet{xu2024dualstream} and the D$^2$-RD-UNet of \citet{cavicchioli2025} explicitly target the classification of interlaced hepatic and portal vein branches rather than producing binary vessel masks, complementing their respective dataset contributions discussed in Section~\ref{subsec:rw_datasets}. Second, graph-based connectivity preservation has gained prominence; the TTGA U-Net \citep{zhao2025ttga}, originally developed for MR-based hepatic vessel segmentation, integrates a two-stream graph attention mechanism into a refinement pipeline, constructing graph structures from superpixel-segmented vessel regions and applying attention-based message passing to propagate connectivity information. Third, label-efficient learning addresses the prohibitive cost of liver vessel annotation: 3D-SLD \citep{10547243} transfers knowledge from 2D structure-agnostic annotations of non-hepatic vascular beds (e.g., retinal and coronary vessels) to 3D liver vessel segmentation via an adversarial shape constraint, while vesselFM \citep{wittmann2025vesselfm} explores foundation-model approaches, training on large-scale multi-organ vascular data to enable few-shot transfer to liver vessels.

\begin{table*}[!t]
\centering
\caption{Overview of liver vessel segmentation methods. Legend -- Metrics: D = DSC, H = Hausdorff, S = Sensitivity, Pr = Precision, cD = clDice, N = NSD, Sp = Specificity, Acc = Accuracy, VOE = Volumetric Overlap Error, BD/TD = Branches/Tree-length Detected. $^*$Post-hoc HV/PV identification via distance voting (not multi-class segmentation). Data: I = 3D-IRCADb, M = MSD Task~08, X = Xu re-annotated MSD, A = AIMS-HPV-385, P = Private, O = Other (non-liver-vessel datasets). Conn. = connectivity-aware architecture or explicit topological enforcement (architectural design, not loss function alone; models using clDice solely as a training loss are not marked).}
\label{tab:rw_methods}
\resizebox{\textwidth}{!}{%
\begin{tabular}{@{}llllp{2.2cm}lllp{3.5cm}@{}}
\toprule
\textbf{Method} & \textbf{Year} & \textbf{Cat.} & \textbf{Architecture} & \textbf{Metrics} & \textbf{Multi} & \textbf{Conn.} & \textbf{Data} & \textbf{Key Contribution} \\
\midrule
\multicolumn{9}{@{}l}{\textit{Classical \& Hybrid}} \\
\addlinespace
\citet{esneault2009liver}       & 2010 & Classical  & Geom.\ moments + graph cuts        & D                & \texttimes & \texttimes & P       & Tubular detection via moments \\
\citet{zeng2017}           & 2017 & Classical  & Oriented flux sym.\ + graph cuts   & S, Sp, Acc       & $\checkmark^*$ & \texttimes & P       & Two-stage coarse-to-fine \\
\citet{sangsefidi2018}     & 2018 & Classical  & Balanced graph cuts                & D                & \texttimes & \texttimes & P, I    & Re-weighted caps for thin vessels \\
\citet{garret2024deep}         & 2024 & Hybrid     & Vesselness fusion + CNN            & D, cD            & \texttimes & \texttimes & I       & Classical + learned features \\
\citet{alirr2023}          & 2023 & Hybrid     & CED/vesselness + U-Net             & D, S, Sp         & \texttimes & \texttimes & M       & Preproc.\ effect on DL \\
\addlinespace
\multicolumn{9}{@{}l}{\textit{CNN-based}} \\
\addlinespace
\citet{kitrungrotsakul2019} & 2019 & CNN       & Multi-pathway 2D CNNs              & D, S, Pr, VOE    & \texttimes & \texttimes & I       & Multi-view extraction \\
\citet{huang2018}          & 2018 & CNN        & 3D U-Net + variant Dice            & D, S, Sp, Acc    & \texttimes & \texttimes & I       & FG--BG imbalance handling \\
HPM-Net \citep{hao2022}    & 2022 & CNN        & Hier.\ progr.\ multi-scale         & D, S, Acc, Sp    & \texttimes & \texttimes & I       & Multi-scale receptive fields \\
LVSNet \citep{yan2021}     & 2021 & CNN        & Attention-guided CNN               & D, S, Pr         & \texttimes & \texttimes & I, P    & Adaptive low-level feat.\ sel. \\
\addlinespace
\multicolumn{9}{@{}l}{\textit{Transformer \& Hybrid CNN--Transformer}} \\
\addlinespace
IBIMHAV \citep{wu2023hepatic} & 2023 & Hybrid  & 3D Swin Transformer                & D, S, Pr, BD, TD & \texttimes & \texttimes & I       & Global context for liver vessels \\
TransRAUNet \citep{transraunet2025}   & 2025 & Hybrid  & TransUNet + Rev.\ Attention        & D, S, Pr, Sp     & \texttimes & \texttimes & I       & Multi-HU windowing augment. \\
HI-Net \citep{liu2024hi}   & 2024 & CNN     & Hier.\ multi-scale fusion          & D, S, Acc, Sp    & \texttimes & \texttimes & I       & Local--global feature integration \\
SCAN \citep{zhou2024scan}     & 2024 & CNN     & Seq.\ context-aware assoc.\ net    & D, S, Pr         & \texttimes & \texttimes & P       & Inter-slice context attention \\
GLIMS \citep{yazici2024glims} & 2024 & Hybrid  & Dilated conv + Swin + CS attn      & D, H             & \texttimes & \texttimes & O       & General volumetric segmentation \\
\addlinespace
\multicolumn{9}{@{}l}{\textit{Multi-task \& Vessel-specific}} \\
\addlinespace
VSNet \citep{xu2025vsnet}             & 2025 & Multi-task & Vessel-growing + Global Dual Tr.  & D, Pr, S, H & \checkmark & \checkmark & X, I    & Centerline regr.\ aux.\ task \\
Xu Dual-stream \citep{xu2024dualstream} & 2024 & Hybrid   & Dual-stream conv--Transf.          & D, S, Pr    & \checkmark & \texttimes & X       & Periph.\ HV/PV misclass.\ red. \\
Li coarse-fine \citep{li2024automated}   & 2024 & CNN        & Coarse-to-fine DL pipeline         & D, H, N, S, Pr & \checkmark & \texttimes & P   & Separate HV/PV/IVC (0.86--0.94) \\
D$^2$-RD-UNet \citep{cavicchioli2025} & 2025 & CNN       & Dual-stage dual-class U-Net        & D, S, Pr    & \checkmark & \checkmark & A       & Connectivity via radius branch. \\
TTGA U-Net \citep{zhao2025ttga}       & 2025 & Hybrid    & U-Net + 2-stream graph attn        & D, cD       & \texttimes & \checkmark & P, I    & Graph attn for vessel topology \\
\addlinespace
\multicolumn{9}{@{}l}{\textit{Label-efficient \& Foundation Models}} \\
\addlinespace
3D-SLD \citep{10547243}     & 2024 & Label-eff.  & 2D$\rightarrow$3D adversarial shape & D, Pr, S  & \texttimes & \texttimes & I      & Vol.$\rightarrow$per-slice annotation \\
vesselFM \citep{wittmann2025vesselfm}  & 2025 & Foundation  & Multi-organ vasc.\ pretrain         & D, cD     & \texttimes & \texttimes & Multi  & Few-shot transfer \\
\bottomrule
\end{tabular}%
}
\end{table*}

Table~\ref{tab:rw_methods} provides a comprehensive overview of all methods discussed in Sections~\ref{subsec:rw_classical}--\ref{subsec:rw_deep}. Notably, most existing methods rely exclusively on DSC-based evaluation, with only a few recent works incorporating topology-aware metrics such as clDice; boundary-sensitive measures like NSD remain largely absent from the liver vessel segmentation literature.

Despite the substantial architectural diversity reviewed above, most methods have been evaluated on a narrow set of datasets with inconsistent annotation quality, and typically report only one or two overlap-based metrics. This limits the ability to draw meaningful cross-study comparisons and to assess clinically relevant aspects of segmentation quality, such as topological connectivity, boundary precision, and geometric fidelity.

\subsection{Evaluation Metrics for Liver Vessel Segmentation}
\label{subsec:rw_metrics}

The evaluation of liver vessel segmentation has historically centered on overlap-based metrics, most commonly the Dice Similarity Coefficient (DSC) and Intersection over Union (IoU). As shown in Table~\ref{tab:rw_methods}, most studies report only DSC, often supplemented by sensitivity or precision. While these measures provide a convenient summary of volumetric agreement, they are fundamentally limited for tubular structures: a segmentation may achieve a high DSC by correctly capturing large central trunks while entirely missing thin peripheral branches, or it may introduce topological disconnections that would be unacceptable in surgical planning yet incur only a marginal penalty in voxel-wise overlap.

Topology-aware and boundary-sensitive alternatives have been proposed to address specific shortcomings. The clDice~\citep{shit2021} metric measures skeleton overlap to capture topological connectivity independently of vessel width, making it valuable for verifying branching completeness. However, clDice is inherently insensitive to vessel caliber errors; the implications of this limitation and its relationship to the Area and Length metrics proposed in this work are analyzed in detail in Section~\ref{subsec:area_length}. The NSD~\citep{nikolov2021} metric evaluates boundary precision within a tolerance threshold, but does not distinguish whether deviations originate from thickness misestimation or from entirely missing segments, nor does it assess longitudinal coverage. Despite their complementary strengths, adoption of these metrics in liver vessel segmentation remains limited: Table~\ref{tab:rw_methods} shows that only a handful of recent works report clDice~\citep{garret2024deep, zhao2025ttga, wittmann2025vesselfm}, while NSD appears in only one study~\citep{li2024automated}.

More critically, a systematic gap persists in the geometric characterization of segmentation quality. No liver vessel segmentation study to date has reported specific measures for \emph{vessel caliber fidelity}, which evaluates how accurately the cross-sectional area of each vessel is represented, or for \emph{longitudinal spatial completeness}, which assesses whether the entire vascular tree is captured, from the proximal trunks to the distal branches. These aspects are clinically indispensable: accurate estimation of lumen diameter directly influences interventional tool selection and volumetric liver function assessment~\citep{selle2002analysis}, while incomplete longitudinal coverage may lead to missed distal branches that are critical for Couinaud-based segment delineation~\citep{fasel1998segmental}. Existing metrics cannot independently quantify these dimensions: IoU and DSC conflate caliber and length errors into a single scalar, clDice verifies connectivity but does not penalize thickness errors, and NSD captures local boundary deviations without assessing global spatial extent.

Following the problem-aware validation principles of Metrics Reloaded~\citep{maier2024metrics}, we propose a five-metric evaluation framework (clDice, IoU, NSD, Area, Length) in which each metric targets a distinct and clinically relevant aspect of vessel segmentation quality. The Area and Length metrics are specifically designed to address the identified geometric gap by quantifying caliber fidelity and longitudinal completeness, respectively, through skeleton-guided analysis with configurable dilation tolerances. Formal definitions, comparative illustrations, and a detailed discussion of how these metrics that complement one another are provided in Section~\ref{evaluationmetric}.

A recent narrative review on AI-driven liver vessel segmentation~\citep{chierici2024} corroborates these observations, noting that despite methodological progress, the field still lacks standardized benchmarks and rigorous multi-metric evaluation. This gap is directly addressed by our proposed dataset and evaluation framework.

\section{Dataset Collection, Labeling, and Challenges}
\label{datasetcoll}

This section describes the clinical origin and technical specifications of our dataset, followed by the rigorous multi-expert protocol used to establish the ground truth. We further detail the diverse range of acquisition artifacts and anatomical variations that characterize this cohort, highlighting the inherent complexities of liver vessel segmentation in a clinical pipeline.

\subsection{Dataset Collection and Characteristics}
\label{collection}

Our dataset was collected from the Radiology Department’s Picture Archiving and Communication System (PACS) at Dokuz Eylül University. The dataset contains 40 abdominal CTA volumes, consisting of 12-bit DICOM images with 512 $\times$ 512 resolution. 20 of these datasets are reserved for training, while the remaining 20 are used for testing. Both datasets (i.e., training and test) contain image series from VEELA's predecessor grand challenge, namely CHAOS \citep{kavur2021chaos}. The number of axial slices in each examination varies between 78 and 294, with 160 slices on average. The $x$- and $y$-spacings are both between 0.54 and 0.79 mm/voxel. Slice thickness varies from 2 to 3.2 mm. These CTA scans belong to 22 female and 18 male potential liver donors. The patient ages ranged from 18 to 57, with an average of 37 years.

The CTA volumes were acquired at the portal venous phase after contrast agent injection. This phase is widely used for liver and vessel segmentation. In this phase, the liver parenchyma is maximally enhanced by the portal veins, but some enhancements also exist for hepatic veins. 

\subsection{Acquisition-related Challenges}
\label{challenges}

\begin{figure}[t!]
    \centering
    \subfloat[\label{fig:fig1a}]{\includegraphics[width=0.485\columnwidth]{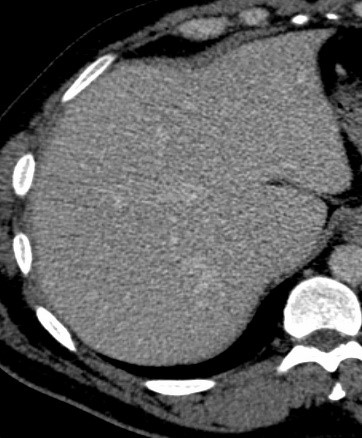}}
    \hfill
    \subfloat[\label{fig:fig1b}]{\includegraphics[width=0.507\columnwidth]{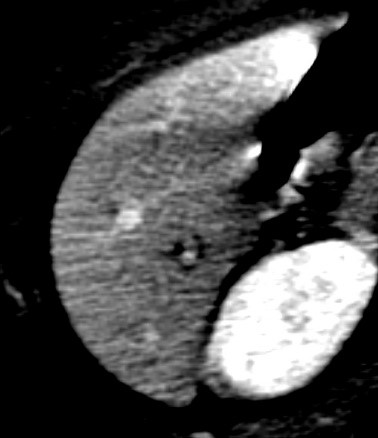}}
    
    \vspace{1em} 
    
    \subfloat[\label{fig:fig1c}]{\includegraphics[width=0.53\columnwidth]{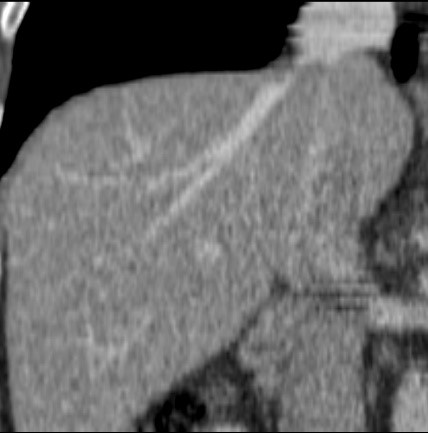}}
    \hfill
    \subfloat[\label{fig:fig1d}]{\includegraphics[width=0.461\columnwidth]{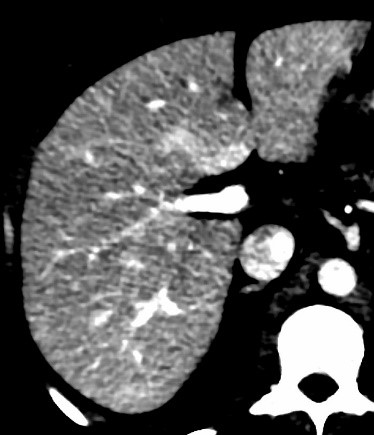}}
    
    \vspace{1em} 
    
    \subfloat[\label{fig:fig1e}]{\includegraphics[width=0.55\columnwidth]{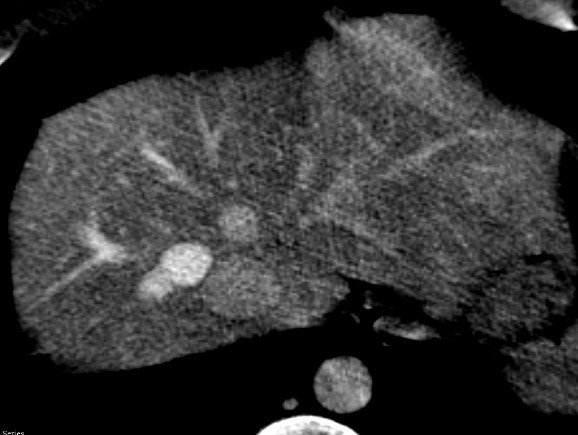}}
    \hfill
    \subfloat[\label{fig:fig1f}]{\includegraphics[width=0.44\columnwidth]{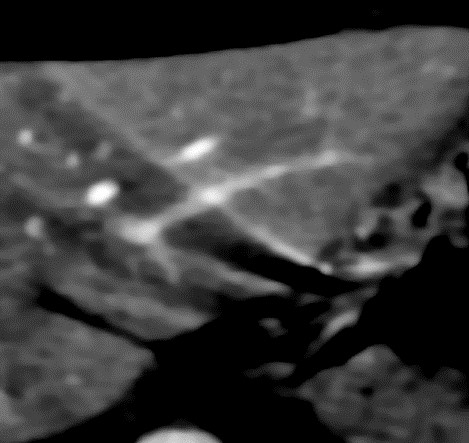}}

     \caption[Examples of quality losses and artifacts.]{Examples of quality losses and artifacts. (a) Quality loss due to miscalculation of timing, (b) artifact causing increased brightness in the upper part of the parenchyma, (c) high inter-slice thickness, (d) transient hepatic attenuation, (e) non-uniform attenuation, (f) beam hardening.}
    \label{fig:fig1}
\end{figure}

CTA acquisitions show high variability in vessel image quality and contrast due to variations in imaging time and conditions (Fig.\ref{fig:fig1}). We categorize the main sources of this variability as follows:

\begin{enumerate}
\item \textit{The miscalculation of timing}: The time period between the injection of contrast agent and the beginning of the scanning can be miscalculated. Therefore, the contrast agent is not in the desired location within the vessel during scanning, and vessel enhancement is not achieved. As a result, the resulting CTA images are of poor quality, and the vessel borders may not be clearly identifiable. Fig.\ref{fig:fig1}a demonstrates an example of a CTA slice with miscalculated timing, where vessels are almost invisible. 

\item \textit{Artifacts}: Artifacts and noise might further drop the quality of rendering. Hepatic and portal vessels may look connected, some parts of the parenchyma may appear brighter and look like vessels, and noise may obscure the appearance of vessels. As an example, Fig.\ref{fig:fig1}b presents a CTA slice for which the top part of the liver appears brighter than the vessels. 

\item \textit{Interslice distance}: High slice thickness alters the adequate appearance of vessels in 3D. Fig.\ref{fig:fig1}c is the coronal view of a liver that was reconstructed from axial plane images. High interslice distance results in low-quality vessel appearance, which is more pronounced on the left side of the inferior vena cava.

\item \textit{Transient hepatic attenuation}: The liver is the only organ with a dual blood supply. Approximately 70\% of its perfusion comes from the portal vein, and 30\% comes from the hepatic artery \citep{wong2008transient}. Transient hepatic attenuations are areas of enhancement on CTA that result from localized variations in the proportion of hepatic arterial and portal venous blood supply. Fig.\ref{fig:fig1}d demonstrates an example, where transient hepatic attenuation is marked with red arrows. The enhanced area similarly resembles the surrounding vessel tissue, which may cause false vessel segmentation detections.

\item \textit{Non-uniform attenuation}: Non-uniform blood velocity into vessel branches complicates the vascular imaging \citep{murphy2018vascular}. A short acquisition time is preferable to obtain uniform opacification of vessels. For instance, Fig.\ref{fig:fig1}e shows that vessels on the right-hand side of the liver have lower intensity values. This phenomenon may mislead methods that operate mainly on grayscale intensities. 
\end{enumerate}

Beyond angiographic acquisition, computed tomography has general drawbacks and limitations. Physics-based artifacts, such as beam hardening, partial volume effect, and undersampling \citep{chen1999spectrum}, may seriously degrade the imaging quality. Fig.\ref{fig:fig1}f presents a slice with beam hardening caused by CT acquisition. Moreover, patient-based artifacts may arise for several reasons, such as patient movement.

\subsection{Labeling: Manual Extraction of Liver Vascular Trees}
\label{labeling}

\begin{figure}[ht!]
    \centering
    \subfloat[\label{fig:fig2a}]{\includegraphics[width=0.48\columnwidth]{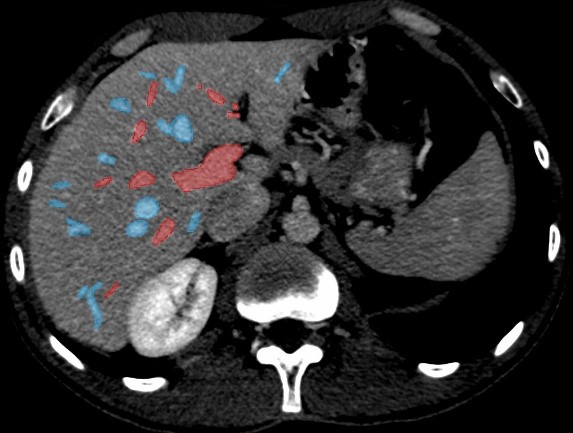}}
    \hfill
    \subfloat[\label{fig:fig2b}]{\includegraphics[width=0.48\columnwidth]{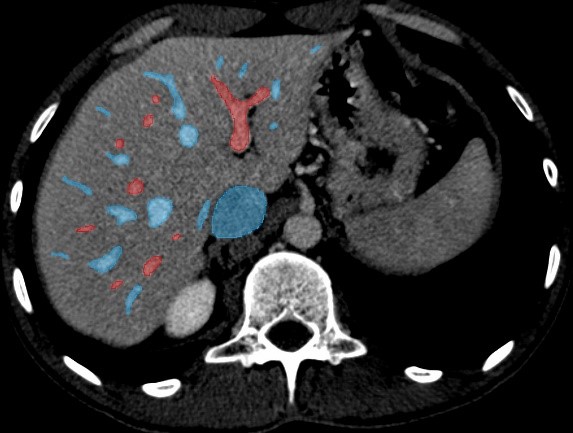}}
    
    \vspace{1em} 
    
    \subfloat[\label{fig:fig2c}]{\includegraphics[width=0.48\columnwidth]{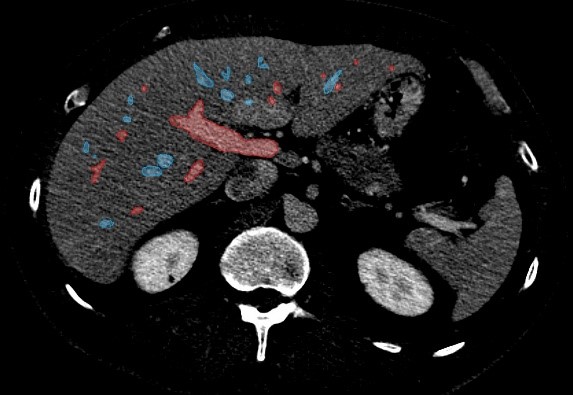}}
    \hfill
    \subfloat[\label{fig:fig2d}]{\includegraphics[width=0.48\columnwidth]{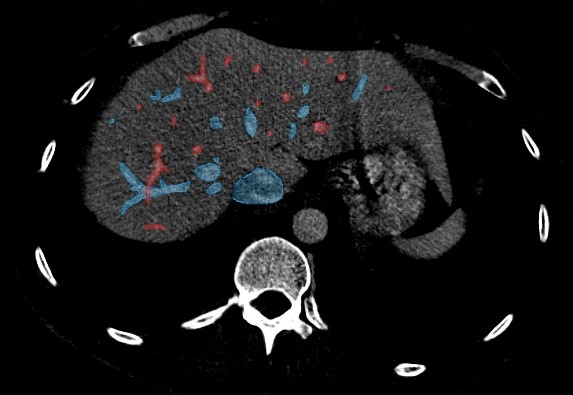}}
    
    \vspace{1em} 
    
    \subfloat[\label{fig:fig2e}]{\includegraphics[width=\columnwidth]{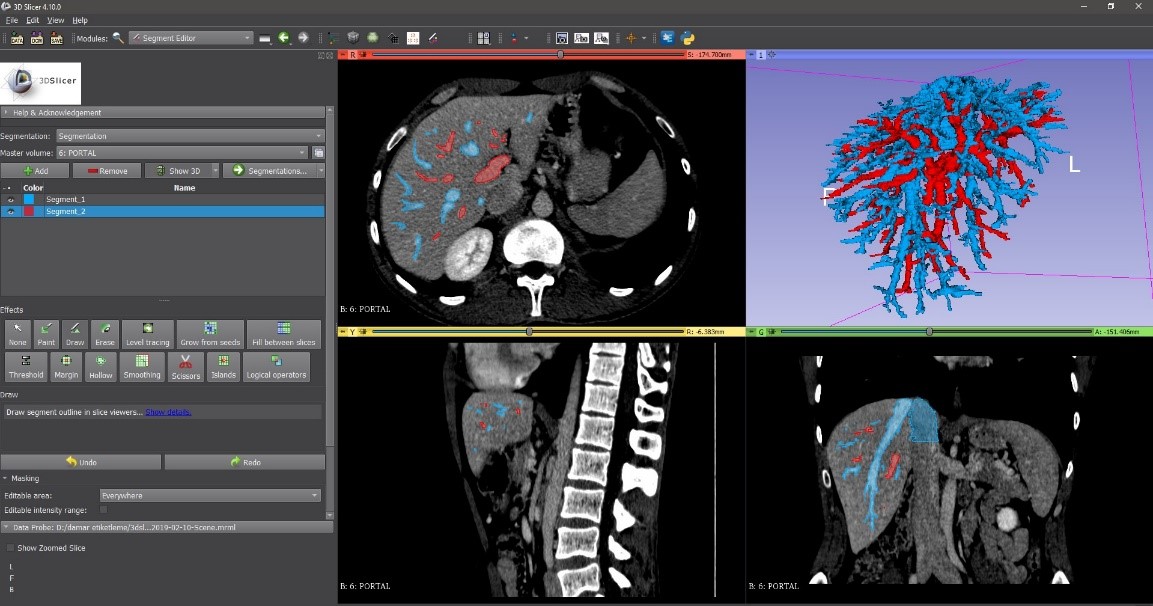}}
    \hfill

    \vspace{1em} 
     
    \subfloat[\label{fig:fig2f}]{\includegraphics[width=0.48\columnwidth]{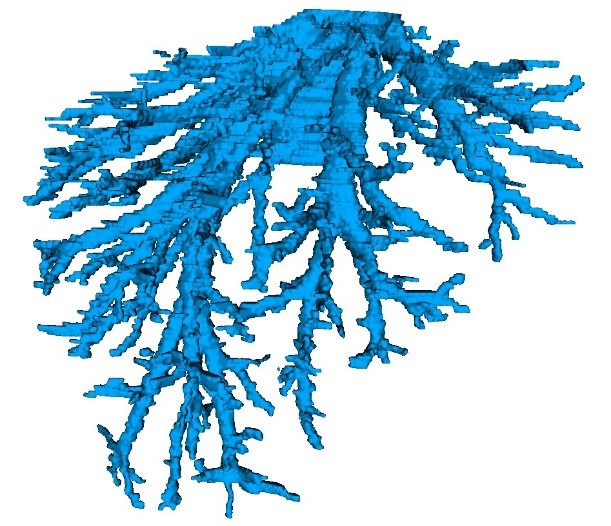}}    
    \subfloat[\label{fig:fig2g}]{\includegraphics[width=0.48\columnwidth]{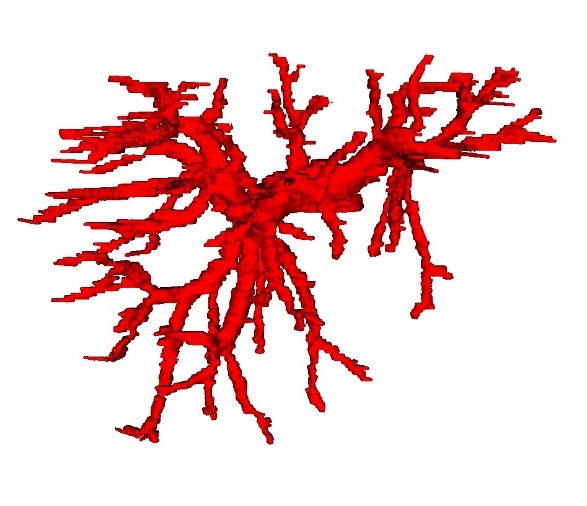}}
    \hfill

    \vspace{1em} 
    
    \subfloat[\label{fig:fig2h}]{\includegraphics[width=0.32\columnwidth]{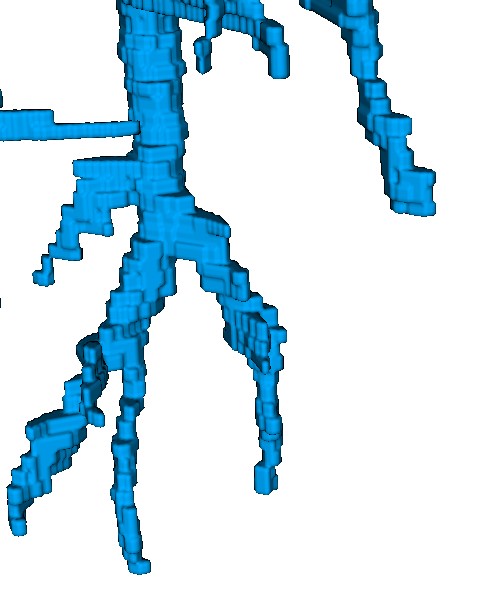}}
    \subfloat[\label{fig:fig2i}]{\includegraphics[width=0.32\columnwidth]{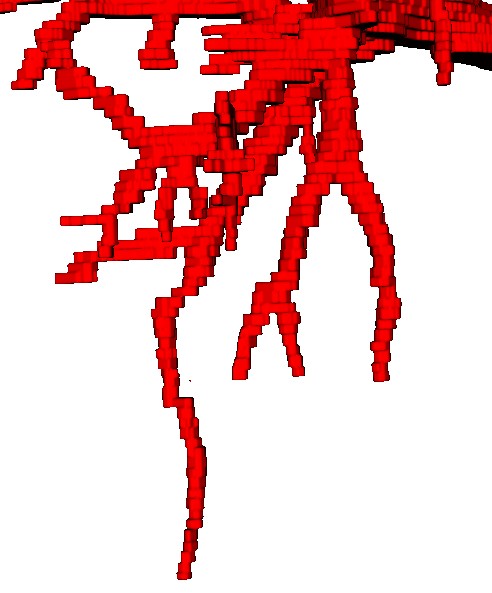}}
      \caption[Examples of labeling]{(a-d) Examples of labeling using axial images. (e) Labeling process using multi-planar reconstructions and visualization in 3D Slicer. (f) Visualization of annotated (g) portal vein (h) hepatic vein. (h-i) Quality control and continuity check of annotated vessels.}
        \label{fig:fig2}
\end{figure}

In this paper, we particularly focused on hepatic and portal vessels. Annotations were performed for all 40 CTA volumes and for each slice (no interpolation was used). An open source program (3D Slicer) \citep{fedorov20123d} was used by a team of two, including an imaging scientist and a radiologist with more than 15 years of experience in portal vessel segmentation. The patient datasets were loaded into the program using the DICOM module, and noise was filtered using curvature anisotropic filtering. By adjusting the window level and width interactively, the vessels were made more visible. Fig.\ref{fig:fig2}a--\ref{fig:fig2}d shows an example labeling process. After labeling all slices, another radiologist with more than 30 years of experience reviewed the vascular tree on 2D slices and generated a 3D model. Thus, a multi-step consensus-based labeling was devised.

\begin{figure}[h!]
    \centering
    \subfloat[\label{fig:fig3a}]{\includegraphics[width=0.47\columnwidth]{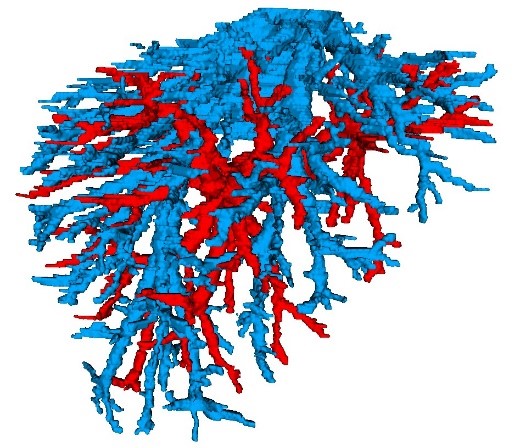}}
    \hfill
    \subfloat[\label{fig:fig3b}]{\includegraphics[width=0.51\columnwidth]{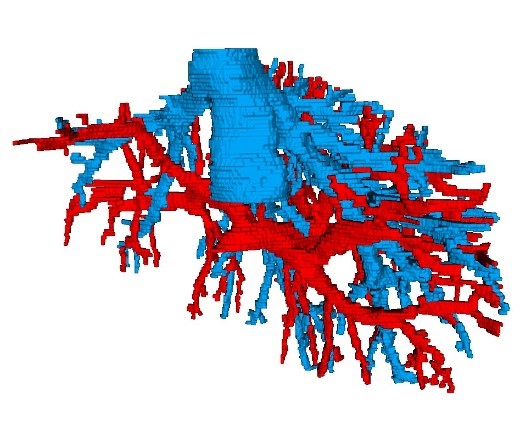}}
    
    \vspace{1em} 
    
    \subfloat[\label{fig:fig3c}]{\includegraphics[width=0.51\columnwidth]{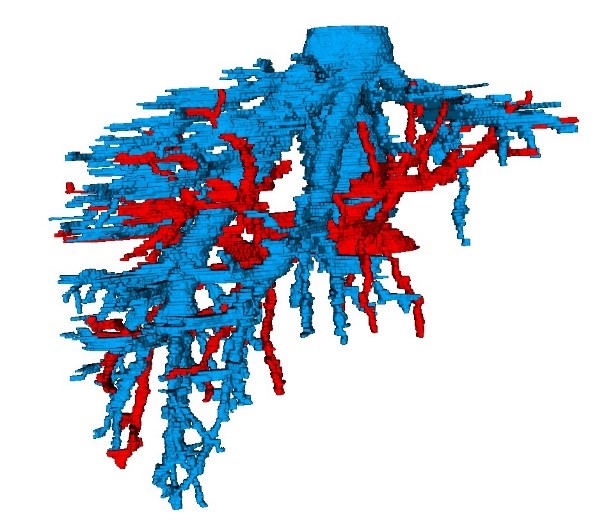}}
    \hfill
    \subfloat[\label{fig:fig3d}]{\includegraphics[width=0.47\columnwidth]{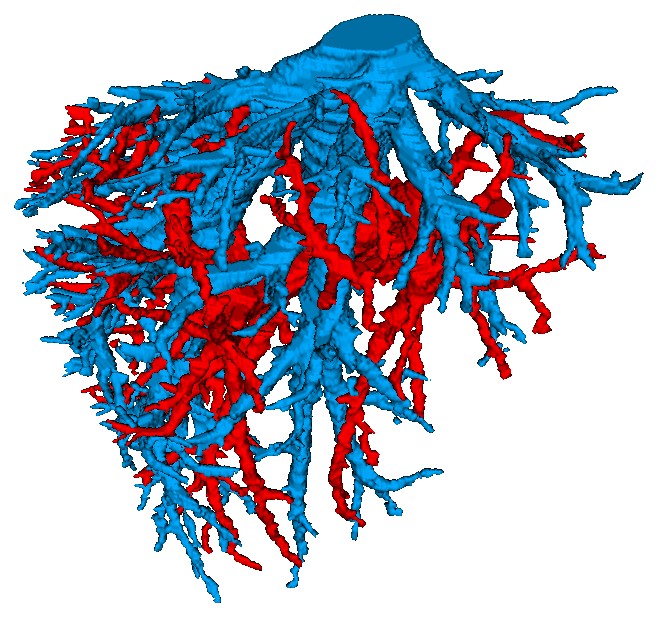}}
    
    \vspace{1em} 
    
    \subfloat[\label{fig:fig3e}]{\includegraphics[width=0.54\columnwidth]{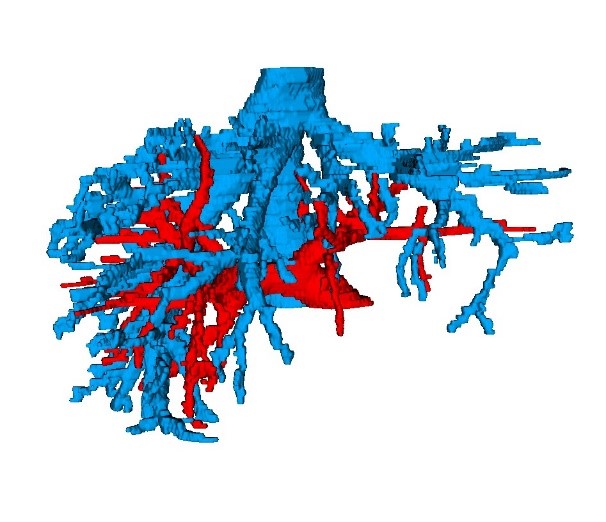}}
    \hfill
    \subfloat[\label{fig:fig3f}]{\includegraphics[width=0.44\columnwidth]{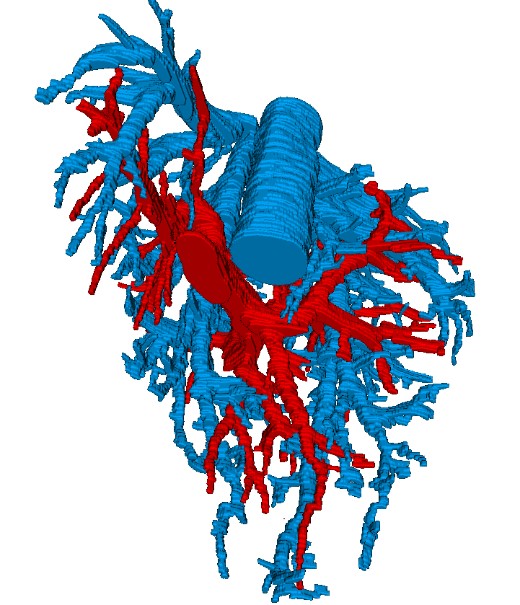}}

      \caption[Examples of 3D models]{Examples of 3D models obtained by directly rendering the labelled data.}
        \label{fig:fig3}

\end{figure}

The program generated 3D models of the labeled vessels, and examining these models allowed us to ensure label continuity. Examples of 3D models of the portal and hepatic veins are shown in Fig.\ref{fig:fig2}f--\ref{fig:fig2}g. Fig.\ref{fig:fig2}h--\ref{fig:fig2}i illustrates zoomed visualizations to allow quality control of annotations considering vessel continuity. All manual segmentations were performed by an expert radiologist with 30 years of experience in the three orthogonal planes. Fig.\ref{fig:fig3} demonstrates examples of visualizations generated by directly rendering the annotated data. 

The manual segmentation process requires a lot of attention; it is very tedious and time-consuming. Nevertheless, it was preferred to create a consistent, rigorous ground-truth image series. Labeling each dataset takes around 8-10 hours, plus the additional time required by another expert for control purposes. 

The main issues encountered during the labeling process were due to unavoidable factors that affected imaging quality (see Section 3.3 for details). Due to routine clinical acquisition standards, some visual degradation occurs at high slice thicknesses. For some datasets, the time elapsed between contrast agent injection and the start of the scan was not properly adjusted. This situation prevented the desired intensity differences between the vessel tissue and the liver parenchyma from being achieved. Imaging quality loss not only prevents the vessel border from being clearly identified, but also causes some visible vessel structures to appear different from their actual size and shape. This makes the images unsuitable as the extracted anatomical structure of the vessels becomes erroneous.

Some CTA artifacts may cause incorrect labeling. Beam hardening is an artifact that occurs when a beam passing through tissues of different densities produces dark lines of varying density \citep{boas2012ct}. It may cause image distortions, leading to misinterpretation of the vessel shape. Transient hepatic attenuation \citep{chen1999spectrum} causes some parts of the parenchyma to be brighter than the remaining tissue, and bright areas may be misinterpreted as vessels. 

\subsection{Examples of Dataset-specific Challenges}

The cases presented in this section illustrate three distinct categories of segmentation complexity encountered in the VEELA dataset. First, imaging-induced visibility constraints arise from acquisition-related factors such as noise, motion artifacts, isodense lumen appearance, and streaking effects, which limit reliable boundary delineation. Second, annotation-policy-induced structural characteristics emerge from VEELA’s visibility-driven labeling principle, leading to intentional discontinuities, non-smooth contours, and selective exclusion of ambiguous regions (e.g., partially intrahepatic IVC segments). Third, true anatomical and post-surgical variations introduce structural deviations from standard vascular patterns, including branching anomalies, altered venous drainage, and resection-related changes. By explicitly distinguishing these categories, VEELA enables a clearer interpretation of segmentation difficulty, separating imaging limitations, annotation design choices, and genuine anatomical variability.

\subsubsection{Variable Intrahepatic IVC Visibility (Set 1):}

\begin{figure}[t!]
    \centering
    \subfloat[]{\includegraphics[width=0.464\columnwidth]{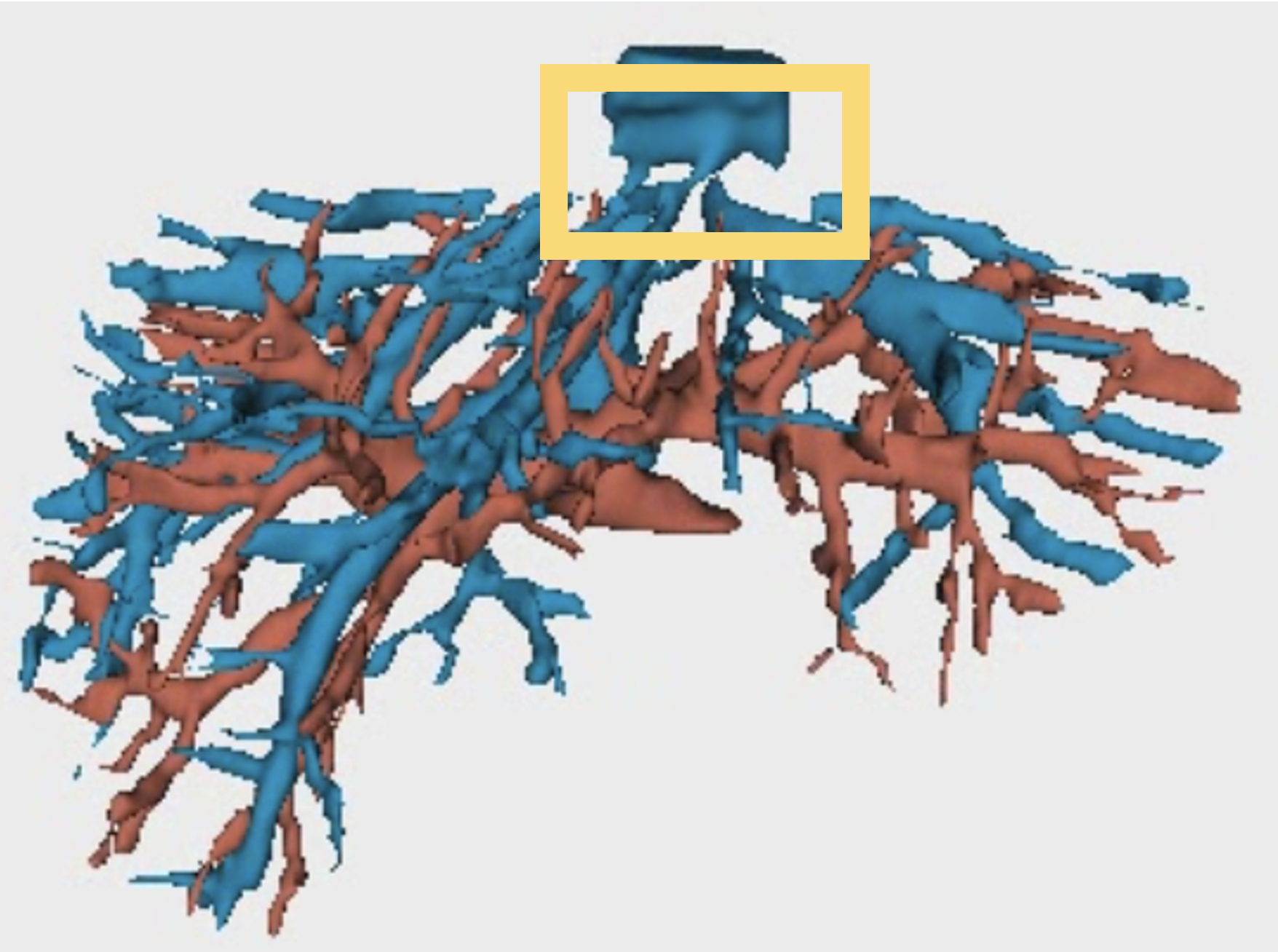}}
    \hfill
    \subfloat[]{\includegraphics[width=0.52\columnwidth]{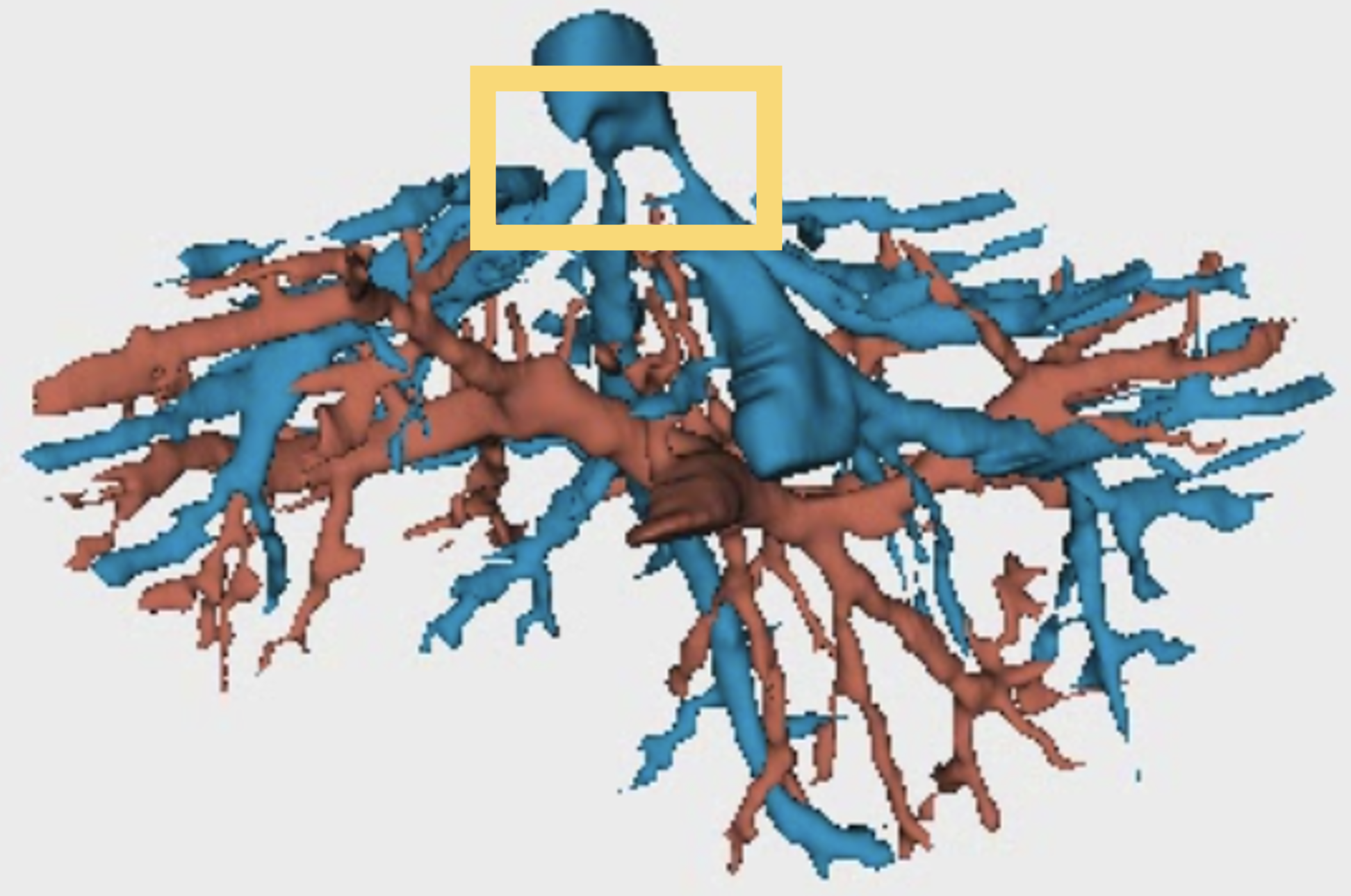}}
    
    \vspace{1em} 
    
    \subfloat[]{\includegraphics[width=0.49\columnwidth]{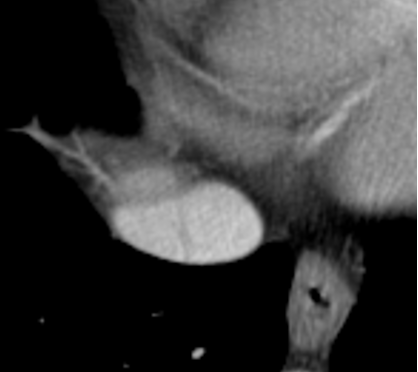}}
    \hfill
    \subfloat[]{\includegraphics[width=0.5\columnwidth]{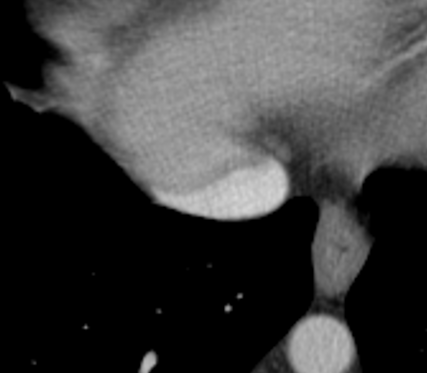}}
    
    \vspace{1em} 
    
    \subfloat[]{\includegraphics[width=0.483\columnwidth]{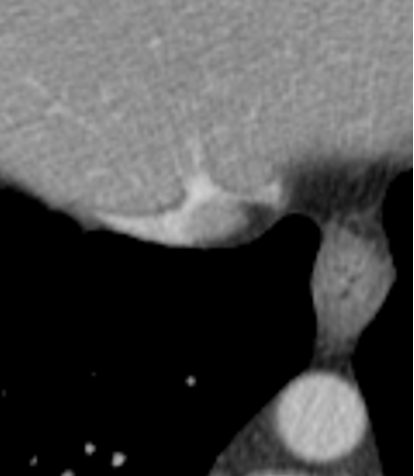}}
    \hfill
    \subfloat[]{\includegraphics[width=0.507\columnwidth]{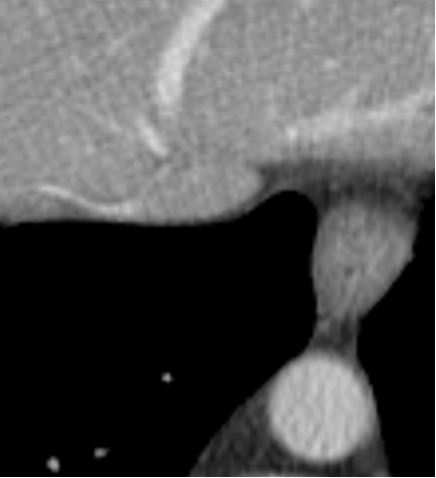}}

     \caption{(a–b) Three-dimensional renderings of the annotated vascular trees (portal veins in red, hepatic veins in blue). The yellow bounding boxes indicate the region of interest corresponding to the intrahepatic IVC. (c–f) Consecutive axial CTA slices through the highlighted region demonstrate variable luminal appearance. The IVC lumen appears well opacified in some slices (c, f) but partially narrowed, compressed, or poorly enhanced in others (d, e). The transition between these appearances occurs without clear anatomical landmarks.}
    \label{fig:ivc_set1}
\end{figure}

The visibility of the vessels varies across consecutive slices. In several regions, the lumen appears partially narrowed, compressed, or inconsistently enhanced. However, it is not possible to determine with certainty whether this apparent narrowing reflects true patient-specific anatomy (e.g., physiological variation or compliance-related deformation) or is due to imaging-related limitations.

This ambiguity highlights a fundamental epistemic limitation of CTA-based vessel delineation: imaging data alone do not reliably distinguish anatomical variation from acquisition artifacts. Similar ambiguities may affect IVC, hepatic, and portal vein branches, particularly in peripheral regions where vessels are thin, faint, or partially obscured.

Because it is impossible to definitively attribute such luminal changes to anatomy or artifact, enforcing anatomically inferred continuity would risk embedding assumptions into the ground truth that are not directly supported by visible evidence. Therefore, VEELA adopts a strict visibility-driven annotation policy across all vascular structures. Vessels are labeled only where lumen boundaries are clearly and consistently identifiable in axial slices. When luminal narrowing, disappearance, or distortion cannot be confidently interpreted, annotation is terminated rather than interpolated.

In this context, the conservative labeling strategy is not an optional design preference but a methodological necessity to ensure that the ground truth reflects observable imaging evidence rather than inferred anatomical expectations. 

The variability illustrated in Fig. \ref{fig:ivc_set1} exemplifies this core challenge in CTA-based vessel delineation. Across consecutive axial slices, the intrahepatic IVC alternates between clearly opacified and partially narrowed appearances, without a definitive indication of whether these changes arise from true anatomical variation or acquisition-related factors. The absence of consistent radiological cues makes it impossible to confidently attribute luminal narrowing to physiological compression, hemodynamic fluctuation, or imaging artifact. In such contexts, enforcing geometric continuity would require inference beyond observable evidence. Therefore, VEELA adopts the aforementioned visibility-driven labeling strategy, which restricts annotation to directly discernible lumen contours. The discontinuities observed in the ground-truth masks should thus be interpreted as reflections of epistemic uncertainty inherent in the imaging process rather than as annotation inconsistencies.

\subsubsection{Selective Exclusion of Extrahepatic IVC (Sets 3 and 4):}

Some portions of the IVC are intentionally left unannotated according to the visibility-driven annotation policy. Although the IVC is frequently visible in portal venous phase CTA, it is excluded from labeling unless it is entirely intrahepatic. 

The IVC often appears adjacent to hepatic veins at the liver boundary, making the distinction between intrahepatic venous structures and extrahepatic caval segments ambiguous on axial slices. Including partially visible or extrahepatic IVC segments would introduce inconsistencies in class definition and potentially bias model learning toward dominant central vessels rather than the targeted hepatic and portal venous systems.

For instance, in sets 3 and 4, IVC regions that extend beyond clearly intrahepatic boundaries are deliberately left unlabeled. This results in abrupt vessel terminations near the hepatic–caval junction, which may visually resemble incomplete annotations but in fact reflect a deliberate and consistent annotation rule.

Importantly, this design choice prevents models from learning anatomically inferred continuity across the liver boundary and avoids over-segmentation of large central vessels. It also ensures that evaluation metrics are not artificially inflated by the inclusion of high-contrast, easily segmentable structures.

\begin{figure}[!t]
    \centering
    \includegraphics[width=\columnwidth]{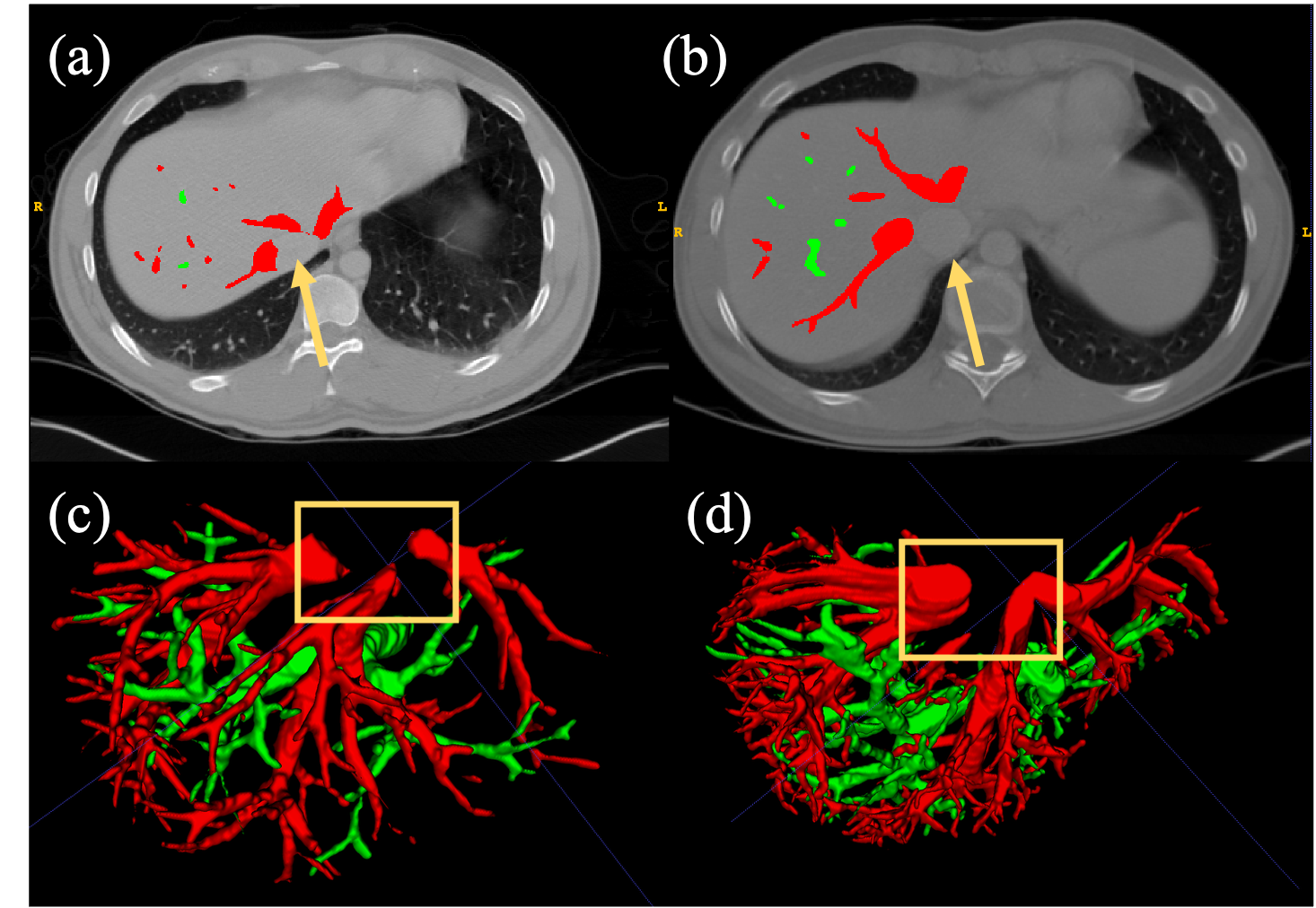}
     \caption{(a–b) Axial CTA slices from Set 3 and Set 4 illustrating portions of the IVC that extend beyond clearly intrahepatic boundaries. Although the lumen remains partially visible, these segments are intentionally left unlabeled under VEELA’s visibility-driven annotation policy. (c–d) Corresponding 3D renderings of the ground-truth vascular trees demonstrate the abrupt termination of the IVC at the hepatic margin. These truncations do not indicate incomplete labeling; they reflect a deliberate scope definition that restricts annotations to intrahepatic venous structures.}
    \label{fig:Set_3_4_VCI_Annot}
\end{figure}

By explicitly encoding the exclusion of partially visible IVC regions, VEELA distinguishes between true segmentation difficulty and ambiguity arising from anatomical scope (Fig. \ref{fig:Set_3_4_VCI_Annot}).

\subsubsection{Isodense Lumen Regions (Sets 3 and 11)}

In certain scans, venous lumens exhibit attenuation values similar to adjacent tissues, particularly in poorly enhanced segments or within portions of the IVC. In these cases, clear boundary delineation is compromised.

\begin{figure}[t]
    \centering
    \includegraphics[width=\columnwidth]{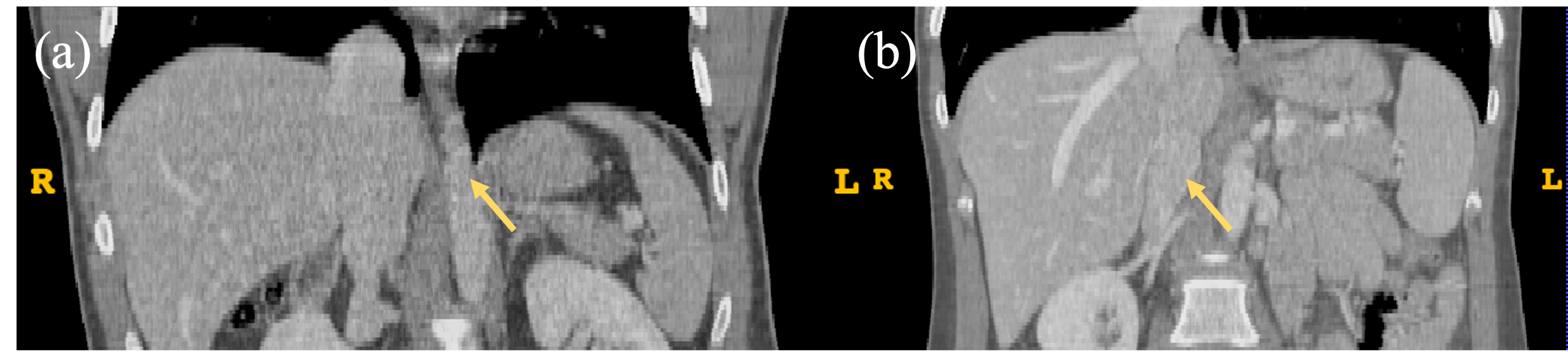}
     \caption{Isodense regions for Set 3 (a) and Set 11 (b) present a challenging situation that complicates the differentiation of the regions.}
    \label{fig:Set3_11_VCI_Isodense}
\end{figure}

Annotations were terminated at the last visible contour. No extrapolation was performed across isodense regions, even when anatomical continuity was expected. This decision preserves the principle that ground truth must reflect observable image evidence rather than inferred geometry (Fig. \ref{fig:Set3_11_VCI_Isodense}).

\subsubsection{Visibility-Constrained Discontinuities and Boundary Irregularities (Sets 3, 7, 11, and 15):}

In these sets, vessel annotations display intentional discontinuities and locally irregular contours. These characteristics arise from imaging-induced visibility constraints rather than annotation inconsistency.

Respiratory motion, noise, isodense regions, and streak artifacts may interrupt lumen visibility across adjacent slices. When boundary confidence is insufficient, annotation is terminated instead of being interpolated. As a result, vascular trees may appear fragmented, particularly in peripheral regions.

Unlike morphology-regularized annotation strategies, VEELA preserves slice-wise boundary appearance without artificial smoothing. This ensures that ground truth reflects imaging evidence rather than anatomically inferred continuity.

\begin{figure*}[t]
    \centering
    \includegraphics[width=\textwidth]{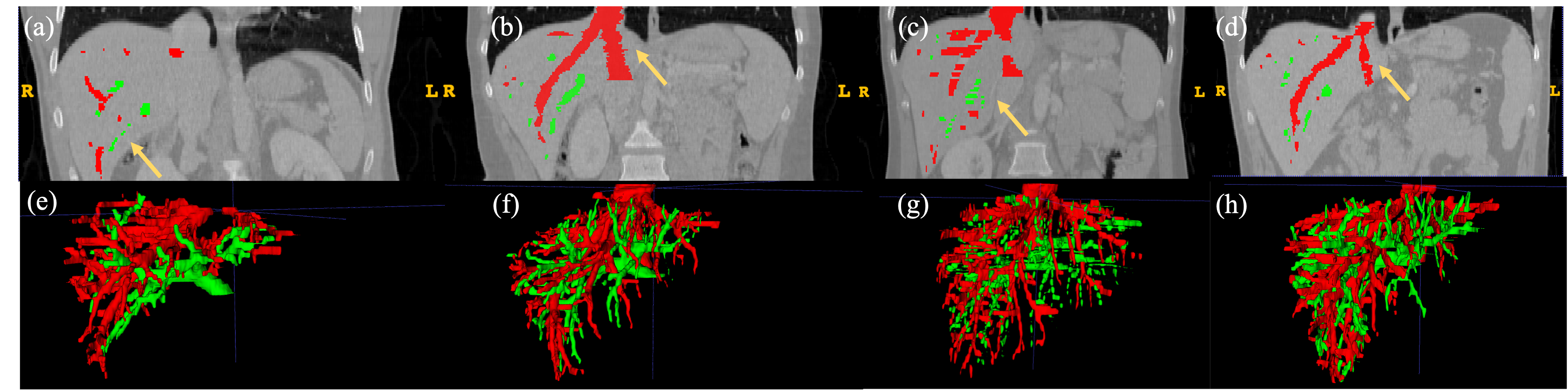}
     \caption{Discrete and non-smooth annotations for some selected coronal image slices from Set 3 (a), Set 7 (b), Set 11 (c), and Set 15 (d) scans. In the bottom row, (e), (f), (g), and (h) represent the 3D ground truth visualizations, respectively.}
    \label{fig:Set3_7_11_15_Discrete_Fleck}
\end{figure*}

Such discontinuities challenge segmentation models to distinguish between true anatomical termination and artifact-induced interruption, making topology-aware metrics particularly informative Fig. \ref{fig:Set3_7_11_15_Discrete_Fleck}.

\subsubsection{Uncertain Visibility Regions / Clinically Indeterminate Vascular Regions (Set 4):}

Specific regions are intentionally left unlabeled due to insufficient visual evidence to reliably delineate vessels. These uncertain visibility regions arise from low contrast, partial volume effects, noise, or motion-related boundary degradation.

When vessel presence cannot be confidently verified in axial slices, no label is assigned—even if anatomical continuity might suggest a trajectory. This conservative strategy encodes the epistemic limits of CTA-based vessel visualization.

\begin{figure}[t]
    \centering
    \includegraphics[width=\columnwidth]{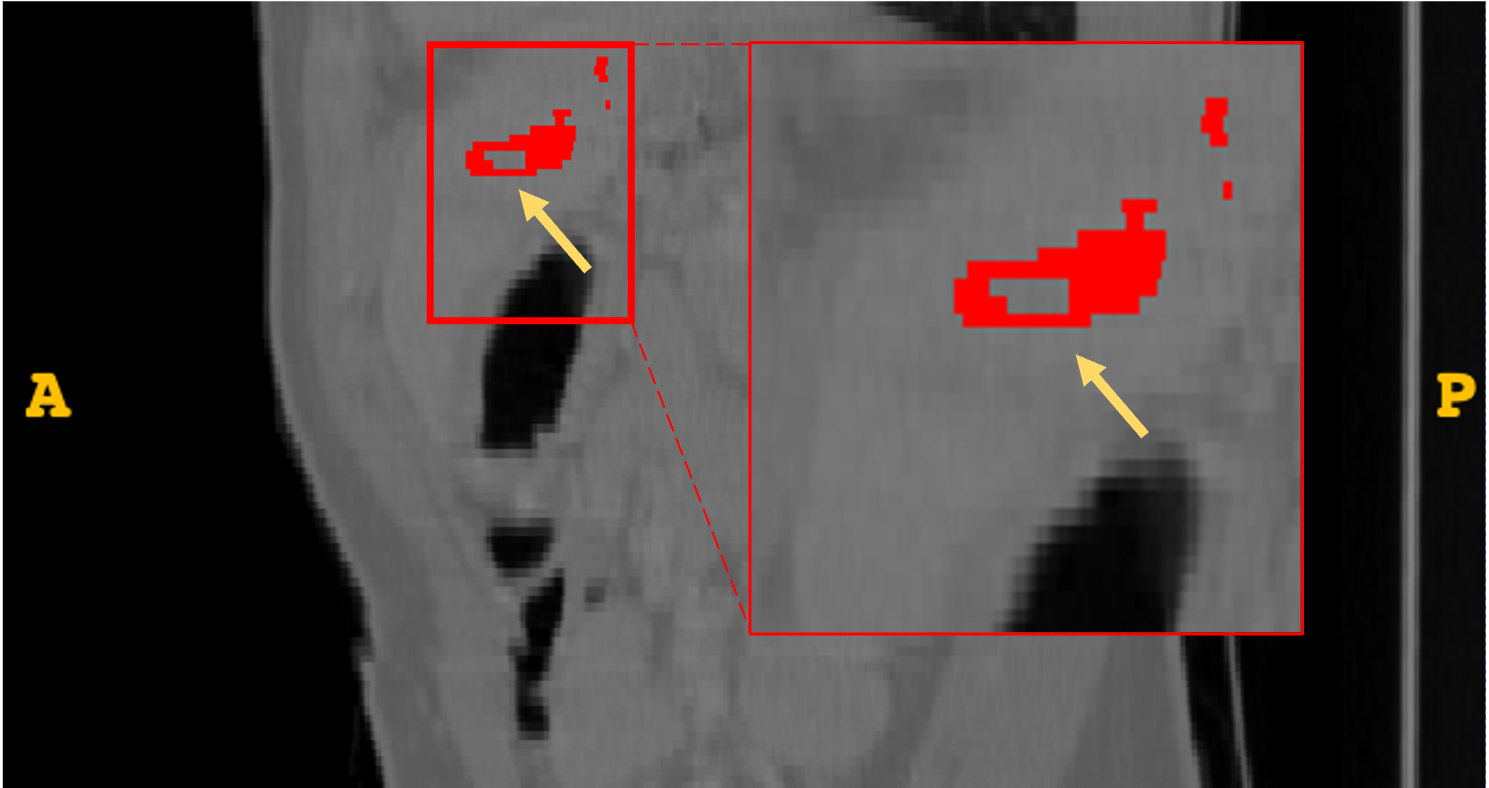}
     \caption{Representative example of a blank region artifact in Set 4. Local signal voids result in a complete loss of image information, leading to the absence of visible vascular structures.}
    \label{fig:Set4_Blank_Region}
\end{figure}

These regions introduce structured ambiguity into the dataset. Models must differentiate between true vessel absence and uncertainty-driven non-annotation, thereby providing a meaningful testbed for evaluating robustness and confidence-aware segmentation strategies (Fig. \ref{fig:Set4_Blank_Region}).

\subsubsection{Perivenous Halo and Pseudovascular Structures (Set 4):}

Bright perivascular halos may mimic additional vascular branches in the axial slices. These appearances are often caused by contrast diffusion, beam hardening, or partial volume effects.

\begin{figure}[t]
    \centering
    \includegraphics[width=\columnwidth]{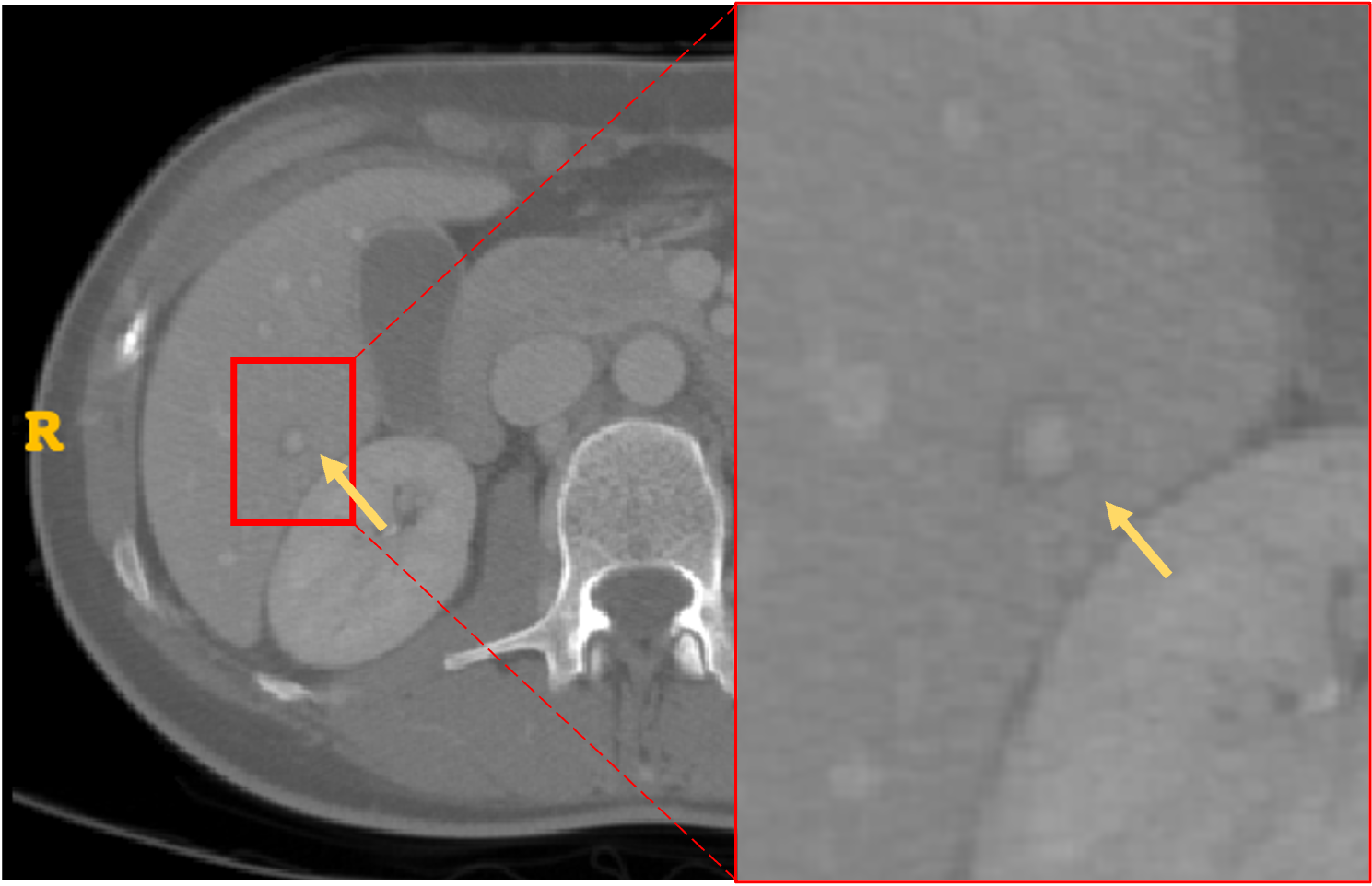}
     \caption{Example of a perivenous space in Set 4. Hypoattenuating regions surrounding vessels may mimic lumen or obscure true vessel boundaries.}
    \label{fig:Set4_Perivenous_Space}
\end{figure}

To avoid mislabeling such regions, annotators required consistent axial traceability before assigning vessel labels. Structures lacking clear continuity across consecutive slices were excluded. This policy reinforces the data set's emphasis on traceable lumen visibility and reduces the risk of artifact-induced false positives (Fig. \ref{fig:Set4_Perivenous_Space}.

\subsubsection{Windmill / Streak Artifacts (Sets 4, 9, and 11): }

Windmill and streak artifacts disrupt vascular continuity across adjacent slices, especially in multi-detector CT acquisitions. These artifacts may obscure lumen boundaries or locally distort vessel shape.

\begin{figure}[!h]
    \centering
    \includegraphics[width=\columnwidth]{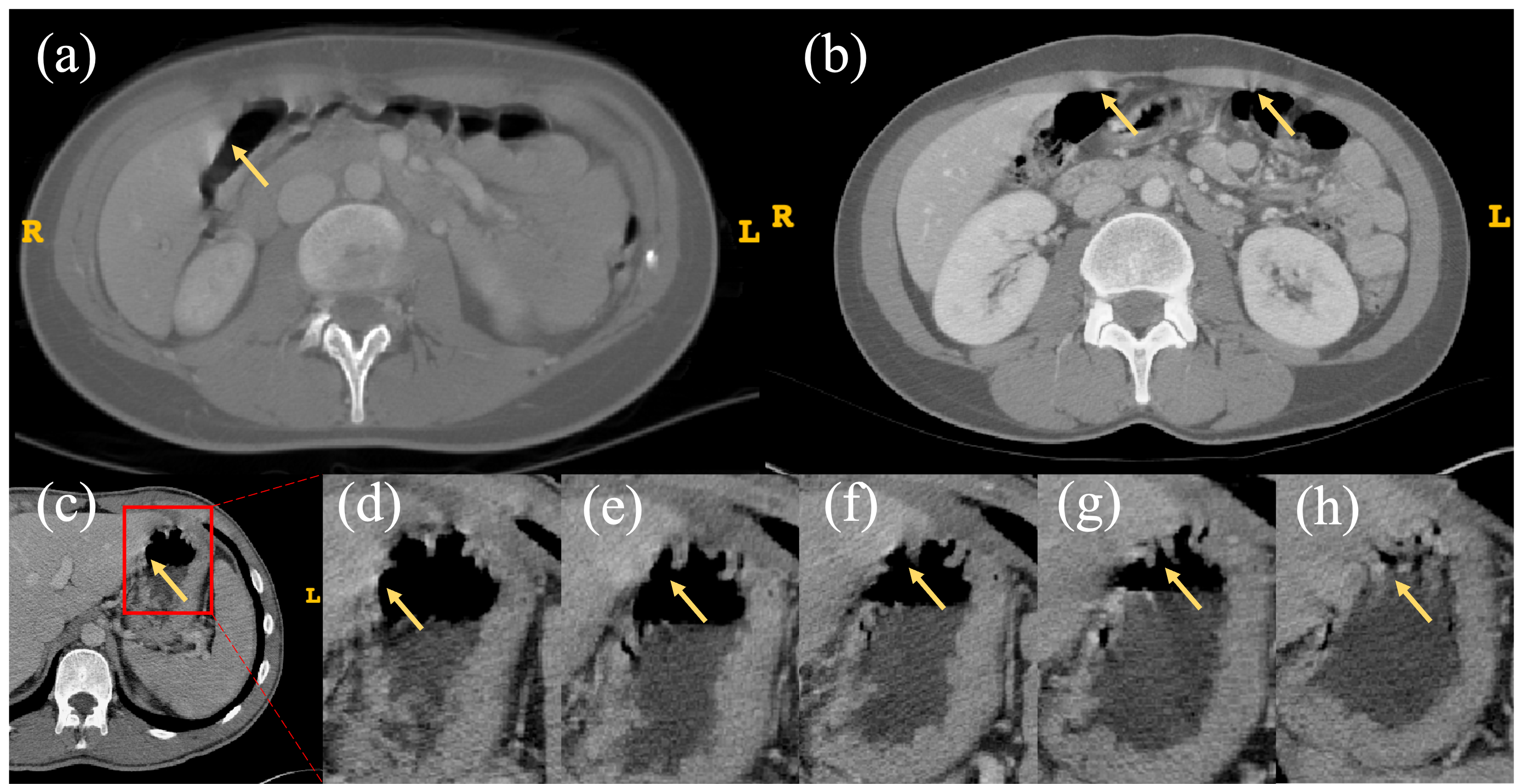}
     \caption{Representative examples of windmill and streak artifacts in Set 4 (a), Set 11 (b), and consecutive slices of Set 9 (c–h). These acquisition-induced artifacts disrupt vessel continuity, obscure lumen boundaries, and introduce structured noise patterns. Annotations are confined to artifact-free regions, resulting in intentional discontinuities. Such conditions evaluate model performance under structured corruption and its ability to preserve topological consistency.}
    \label{fig:Set4_9_11_Windmill}
\end{figure}

In VEELA’s annotation framework, vessels were labeled only in artifact-free regions. Where streaking compromised visibility, annotations were intentionally discontinuous. Such discontinuities do not imply anatomical absence but rather encode acquisition-induced uncertainty in the dataset (Fig. \ref{fig:Set4_9_11_Windmill}).

\subsubsection{Signal-to-Noise Ratio (SNR) Conditions (Sets 7 and 9):}

\begin{figure}[!h]
    \centering
    \includegraphics[width=\columnwidth]{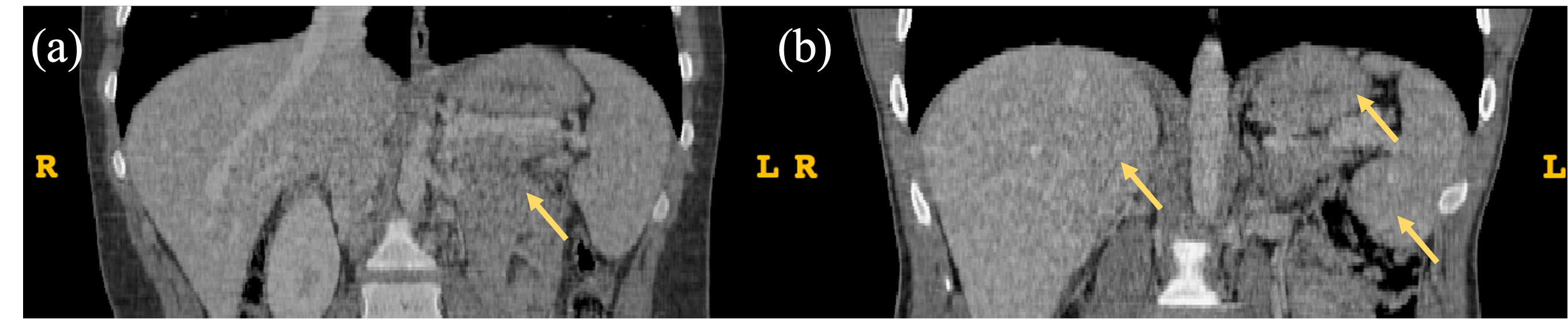}
     \caption{Representative examples of low signal-to-noise ratio (SNR) conditions in Set 7 (a) and Set 9 (b). Elevated noise levels reduce contrast between small-caliber vessels and surrounding liver parenchyma, leading to indistinct vessel boundaries, particularly in deeper regions.}
    \label{fig:Set7_9_Noise}
\end{figure}

High noise levels reduce the contrast between small-caliber vessels and the surrounding parenchyma, particularly in deep liver regions. Under such conditions, the boundaries of the vessel may become indistinct. Annotators refrain from inferring the presence of vessels in ambiguous areas. This ensures that segmentation masks reflect image-supported evidence rather than speculative continuity. Low-SNR conditions evaluate a model’s ability to preserve peripheral vessel structure under global degraded imaging quality (Fig. \ref{fig:Set7_9_Noise}).

\subsubsection{Respiratory Motion Artifacts (Sets 7, 9, and 11):}

Respiratory motion during acquisition introduces slice-wise misalignment and boundary blurring, particularly evident in peripheral branches and multi-planar reconstructions.

When motion compromises the clarity of the lumen boundary, annotation is restricted to slices that are confidently visible. No interpolation is performed across motion-distorted regions.

\begin{figure}[!h]
    \centering
    \includegraphics[width=\columnwidth]{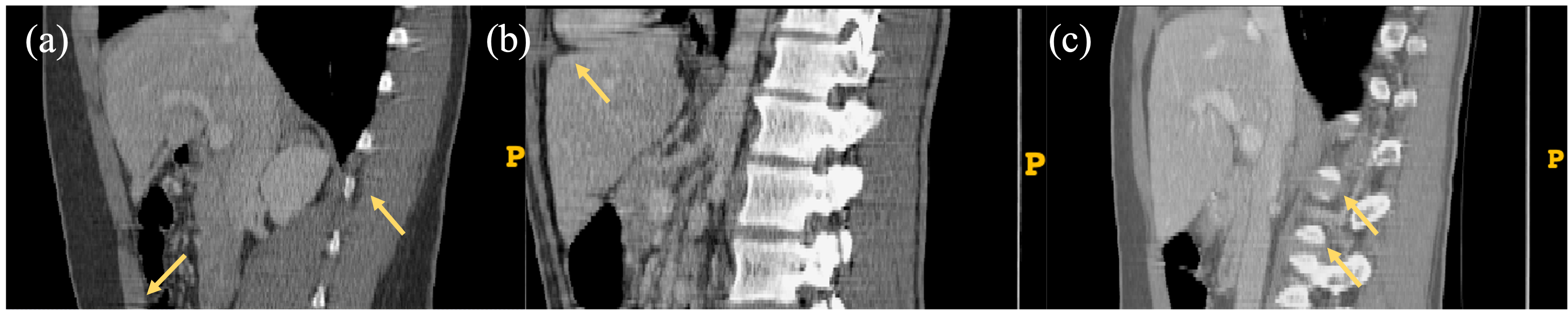}
     \caption{Examples of motion-induced fragmentation and respiratory artifacts. Breathing-related motion introduces slice-wise misalignment and boundary blurring, particularly affecting peripheral branches.}
    \label{fig:Set7_9_11_Breathing_Artifact}
\end{figure}

This produces localized discontinuities that reflect acquisition dynamics rather than anatomical interruption, challenging models to differentiate motion-induced fragmentation from true vascular termination (Fig. \ref{fig:Set7_9_11_Breathing_Artifact}).

\subsubsection{Set 14: Biphasic IVC and Portal Vein Trifurcation}

\paragraph{{Issue 1: The IVC remains bi-phasic}}

\begin{figure}[h!]
    \centering
    \includegraphics[width=\columnwidth]{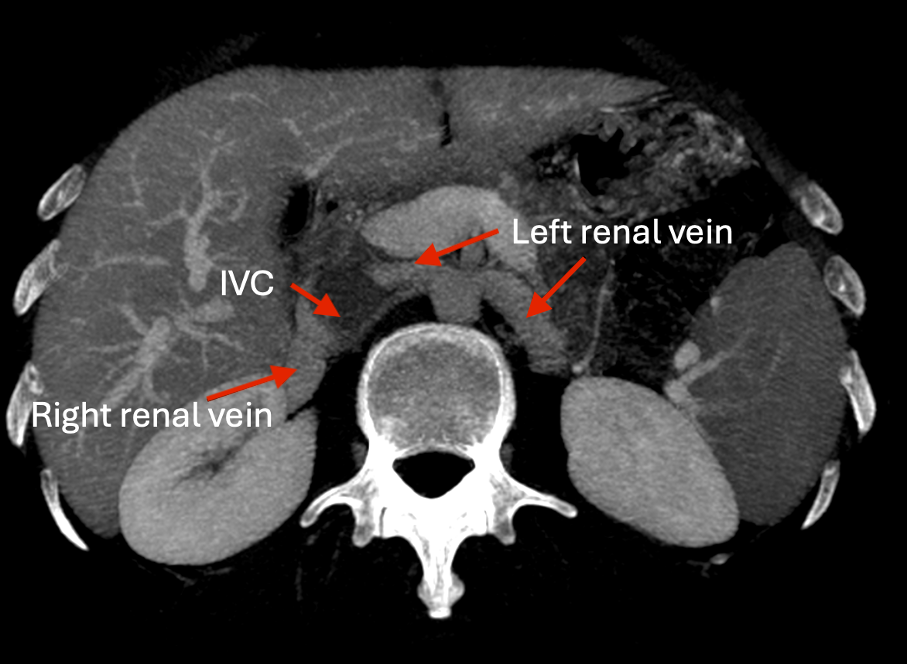}
    \caption{Set 14, Issue 1. Biphasic IVC appearance due to early venous phase acquisition. 
    Enhanced blood from the renal veins and poorly enhanced blood from the lower body 
    did not fully mix.}
    \label{fig:biphasic_ivc}
\end{figure}

The expression "remains biphasic" suggests that, even after contrast agent administration, the IVC did not achieve uniform opacification or show clear enhancement patterns (Fig. \ref{fig:biphasic_ivc}). This may indicate a hemodynamic issue or a problem with contrast agent distribution. Potential causes could include:
\begin{itemize}
    \item \textit{Rapid renal venous return during early venous phase}: This results in more intense enhancement in the periphery of IVC relative to the venous return from the hepatic veins at the early venous phase. This temporary appearance fades by the late venous phase. IVC contrast may appear biphasic if the image acquisition window coincides with the early venous phase.
    \item \textit{Obstruction or narrowing}: There might be stenosis (narrowing) or a thrombus (clot) in the down- or upstream vessels, preventing the contrast agent from reaching that area effectively or delaying the venous return, which results in contrast difference in the intravenous compartment.
    \item \textit{Abnormal blood flow}: Reduced or disrupted blood flow, either due to cardiovascular disturbances or due to systemic problems, may cause the vessel to appear biphasic.
    \item \textit{Vasospasm}: Temporary narrowing of the vessel (vasospasm) due to local or systemic causes, which may affect the autonomic nervous system or humoral feedback mechanisms, could result in this appearance.
    \item \textit{Contrast administration error}: Issues with the injection of the contrast agent, such as improper timing, small caliber intravenous access, or insufficient delivery of the contrast agent, can also cause this phenomenon.
\end{itemize}

\paragraph{{Issue 2: Portal vein branching variation}} 

\begin{figure}[t]
    \centering
    \subfloat[]{\includegraphics[width=0.35\columnwidth]{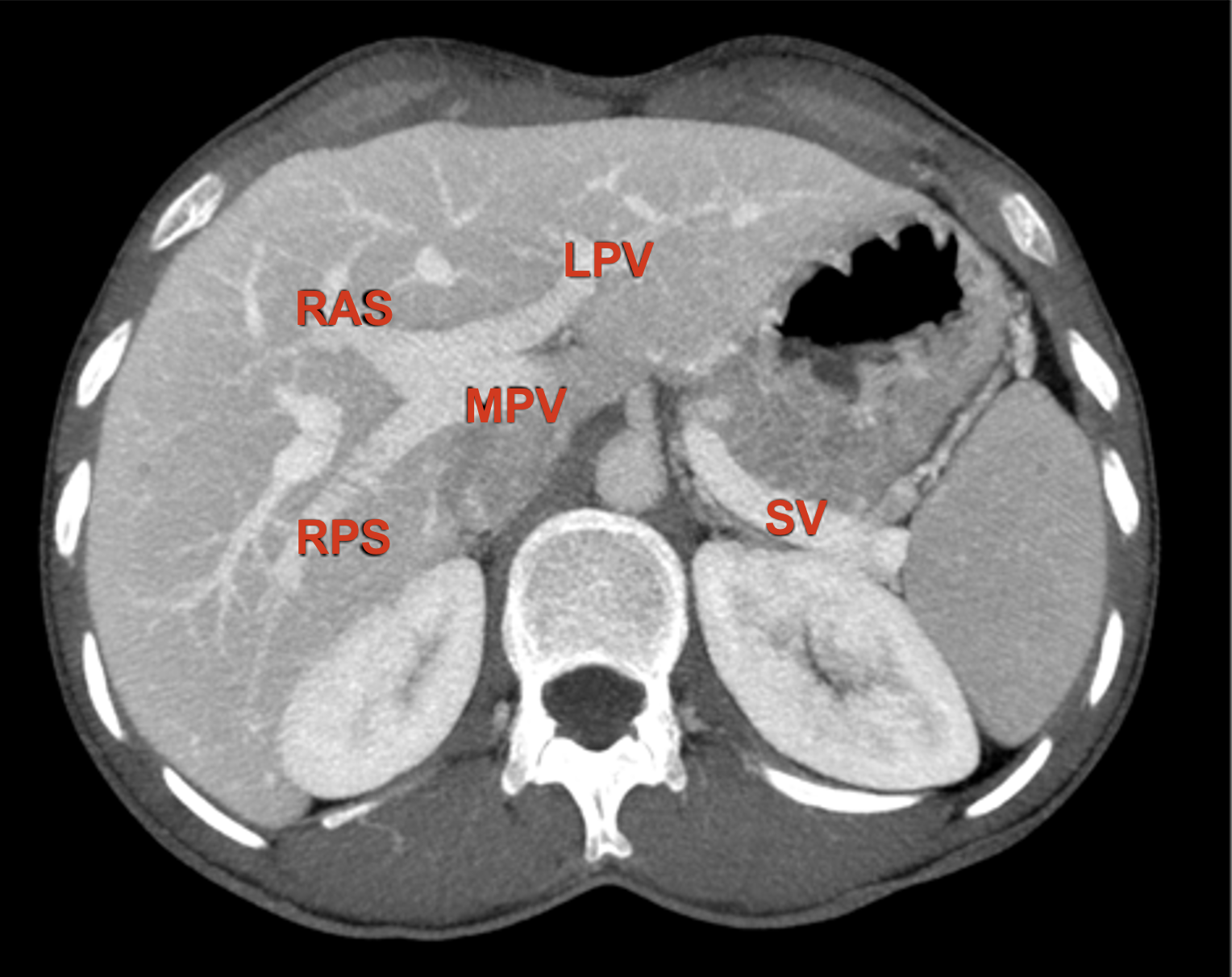}}
    \hfill
    \subfloat[]{\includegraphics[width=0.336\columnwidth]{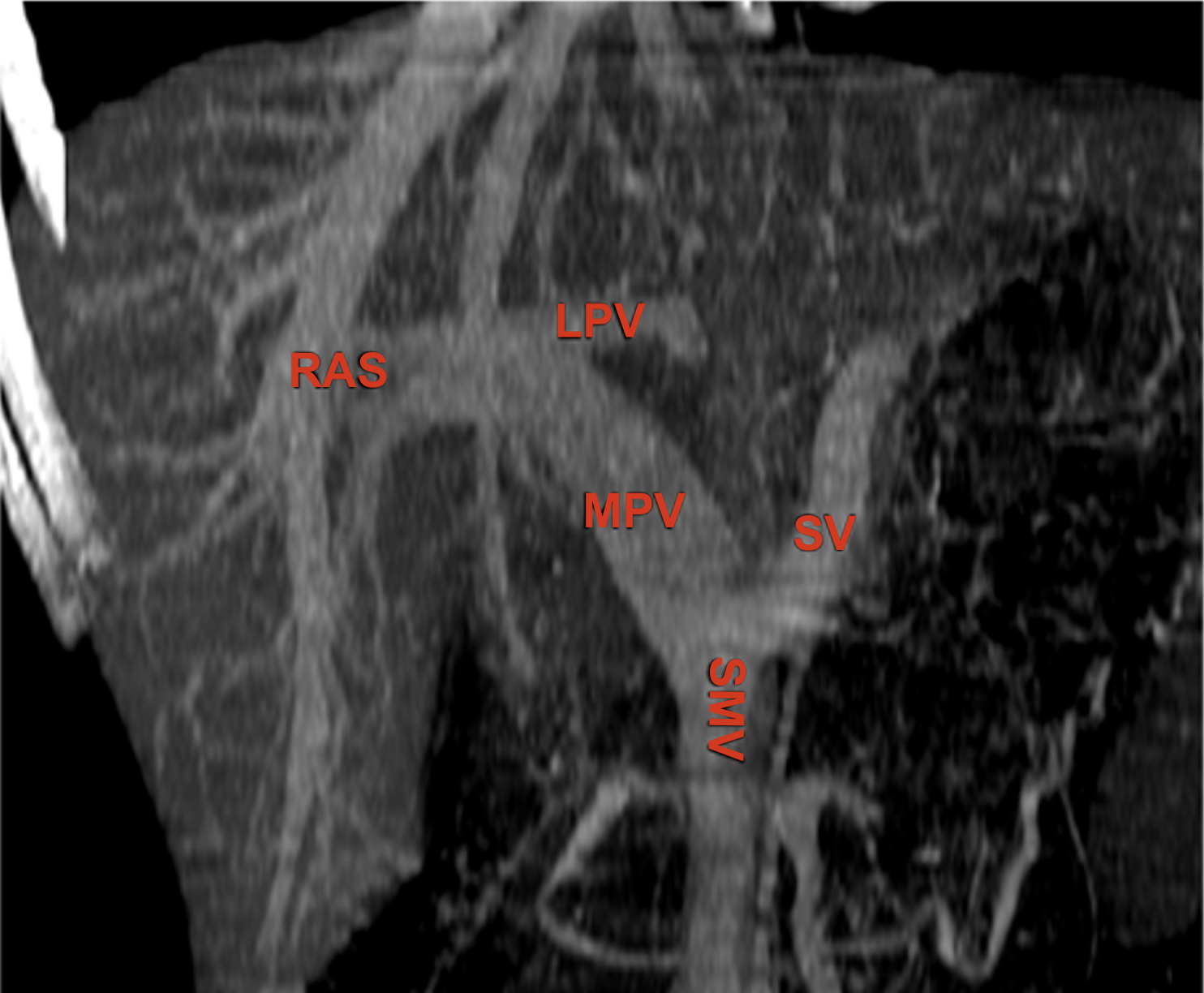}}
   \hfill 
    \subfloat[]{\includegraphics[width=0.303\columnwidth]{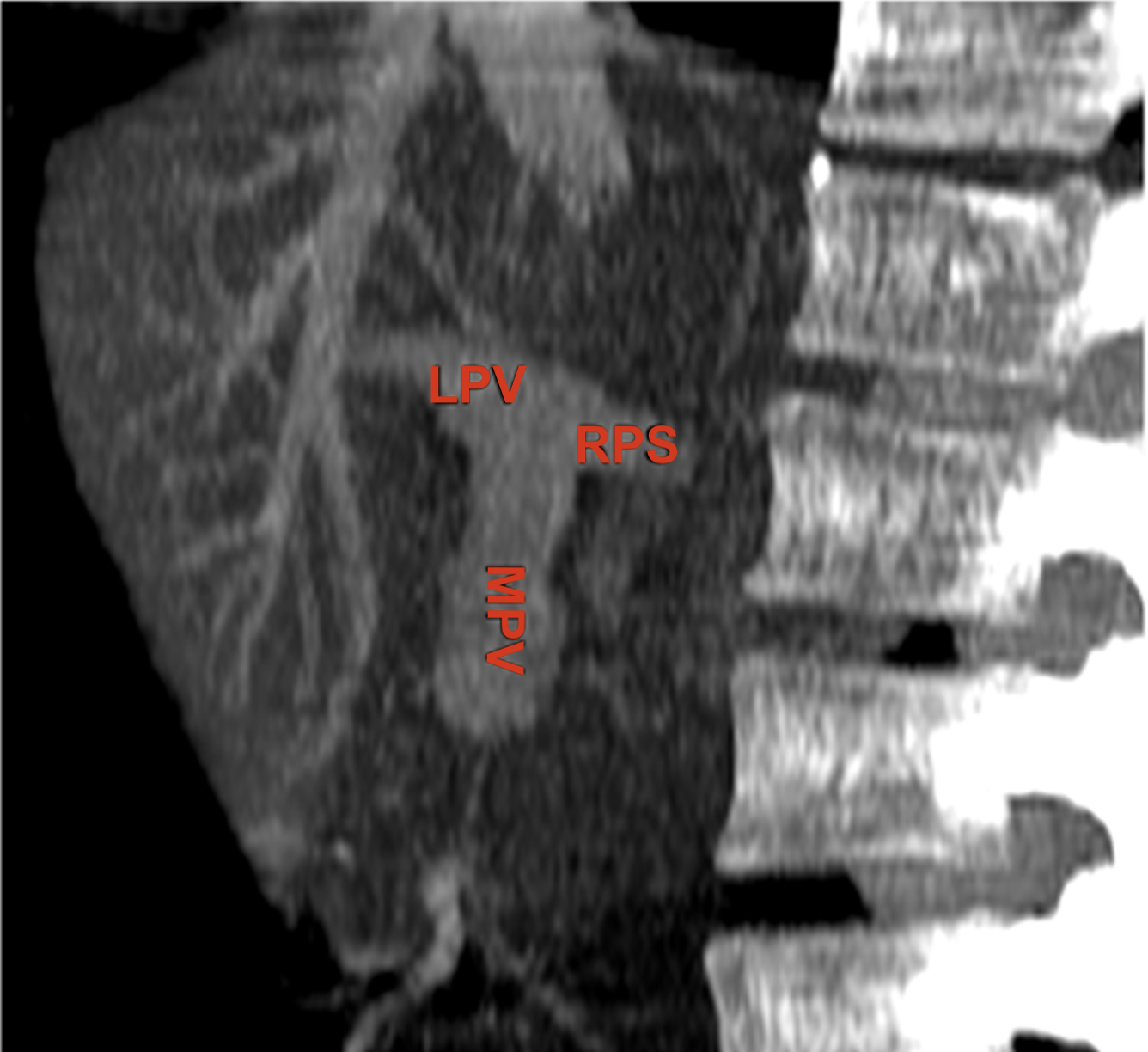}}
    \hfill
    
     \caption{ Set 14 Issue 2: Portal vein trifurcation. Left: Axial oblique thin MPR. Middle: Coronal oblique thick MIP (Maximal Intensity Projection). Right: Sagittal oblique thick MIP reconstruction.}
    \label{fig:portal_vein_trifurcation2}
\end{figure}

There is a portal vein trifurcation, which is an anatomical variation of the portal vein in this patient (Fig. \ref{fig:portal_vein_trifurcation2}). Normally, the portal vein divides into two main branches within the liver: the Right Portal Vein (RPV) and the Left Portal Vein (LPV). In cases of trifurcation, however, the portal vein divides into three branches: 1) Right Anterior Segmental Branch (RAPV - Right Anterior Portal Vein), 2) Right Posterior Segmental Branch (RPPV - Right Posterior Portal Vein), and 3) LPV. The right portal vein then further divides into the anterior and posterior branches. In trifurcation, the right anterior segmental branch, the right posterior segmental branch, and the left portal vein arise simultaneously from the main portal vein trunk. While portal vein anatomical variations are usually asymptomatic, they can become significant in certain situations, including liver surgery. During liver transplantation, resection, or tumor surgery, recognizing this variation is crucial. If trifurcation is present, accidental ligation or damage to one of the branches can lead to serious complications.

Since the orientations of RAS, RPS, and LPV are orthogonal in the axial plane and their presence plane is orthogonal to the MPV orientation, all branching patterns cannot be presented in a single long image plane. Splenic Vein (SV), which drains the spleen, and the Superior Mesenteric Vein (SMV), which drains the gastrointestinal tract from the Treitz ligament to the splenic flexure, unite to form the Main Portal Vein (MPV).

\subsubsection{Set 24: Absence of Hepatic Branches in Segments 5-6}

\begin{figure}[t]
    \centering
    \subfloat[]{\includegraphics[width=0.524\columnwidth]{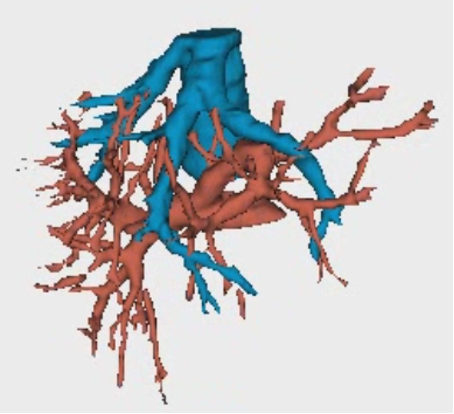}}
    \hfill
    \subfloat[]{\includegraphics[width=0.47\columnwidth]{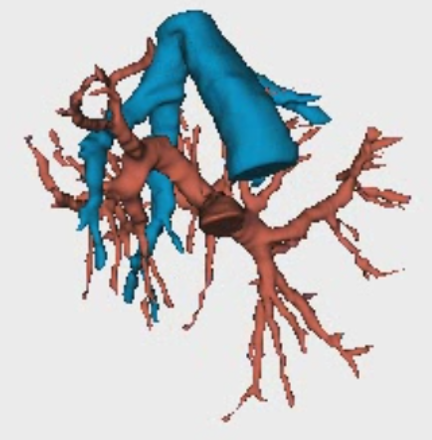}}
    
    \vspace{1em} 
    
    \subfloat[]{\includegraphics[width=0.51\columnwidth]{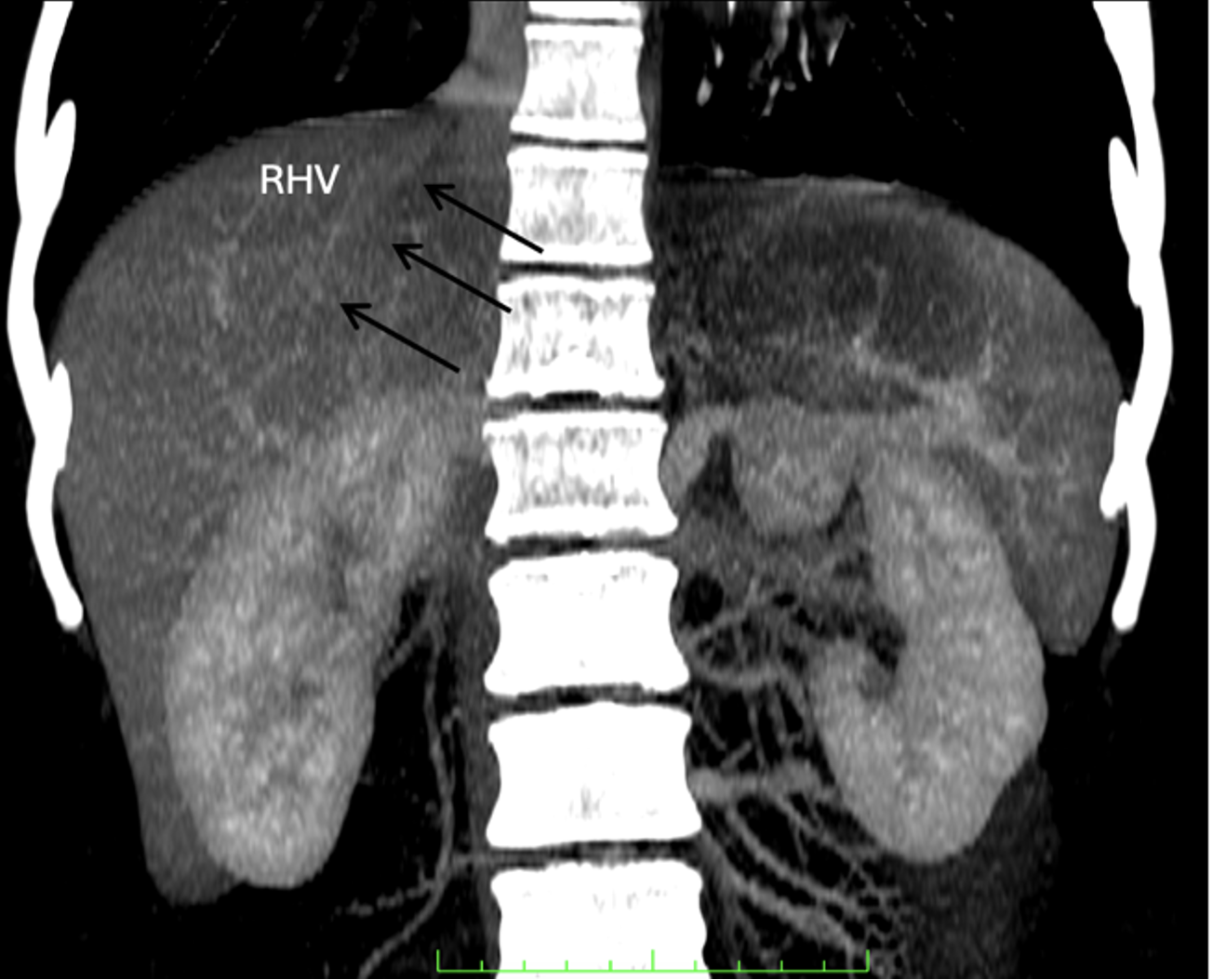}}
    \hfill
    \subfloat[]{\includegraphics[width=0.48\columnwidth]{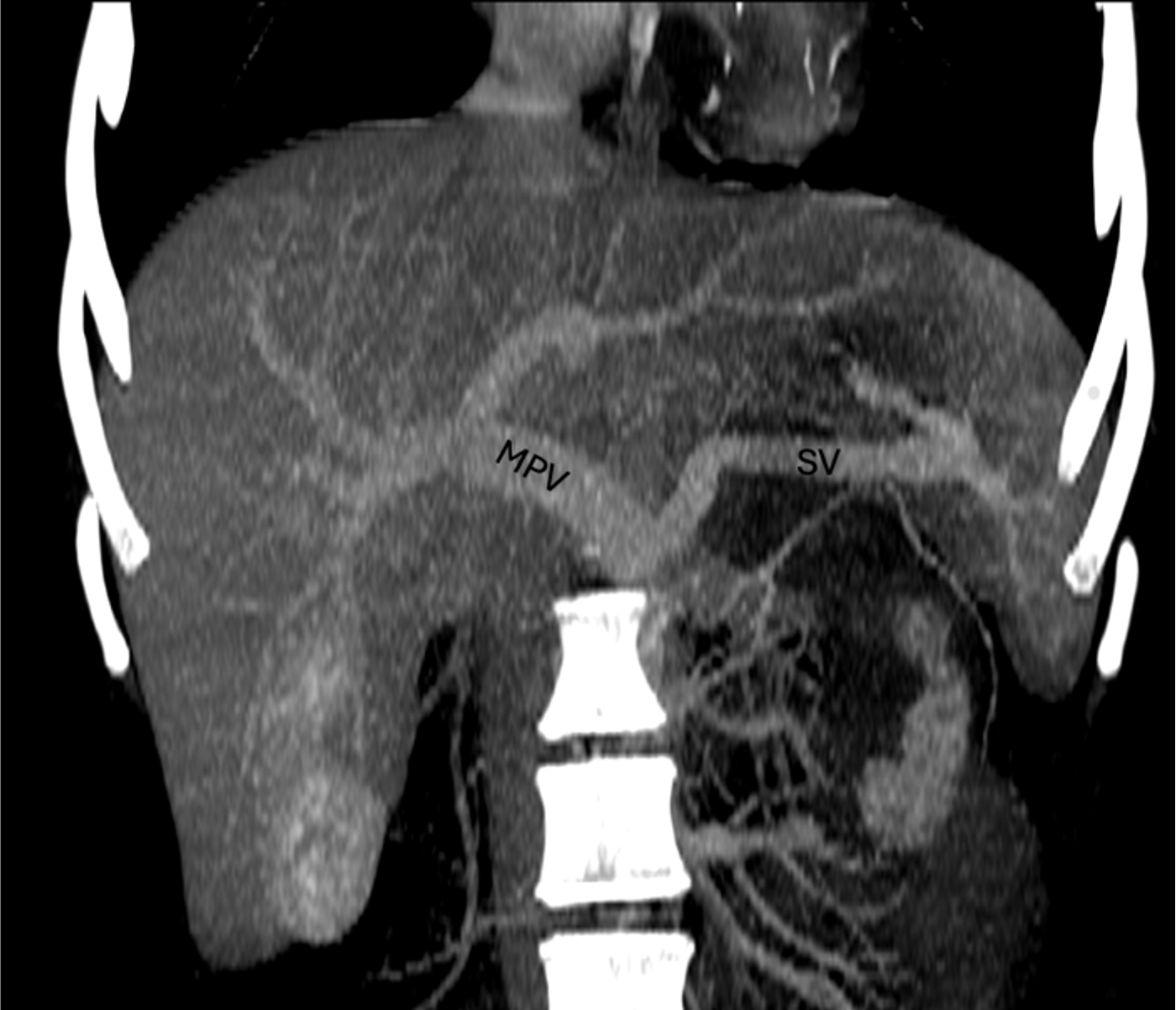}}
        
    \caption{Set 24. Absence of hepatic veins in segments 5 and 6. The right hepatic vein (RHV) 
    is barely visible and cannot be traced below the portal vein (MPV) level. Poor opacification 
    precludes reliable tracing of smaller veins.}
    \label{fig:hepatic_vein_absence}
\end{figure}

This condition refers to a rare anomaly where the normal venous drainage pathways, specifically the branches of the hepatic veins, are absent from certain liver segments (Fig. \ref{fig:hepatic_vein_absence}). This variation may arise from developmental differences in the hepatic venous system. Possible causes for this condition include:

\begin{itemize}
    \item \textit{Developmental anomalies}: During embryological development of the liver's venous system, certain hepatic vein branches may fail to develop or may form with abnormal drainage patterns.
    
    \item \textit{Alternative drainage pathways}: Venous drainage from segments 5 and 6 might bypass the right hepatic vein and drain directly into the Inferior Vena Cava (IVC) or into other hepatic veins.

    \item \textit{Atrophic or Hypoplastic veins}: Hepatic vein branches may be present but too small or hypoplastic to be visualized on imaging studies.
 
    \item \textit{Poor imaging conditions}: Acquisition noise may reduce the visibility of small venous structures. Although this condition is often asymptomatic and discovered incidentally, awareness of it is crucial during surgical or interventional procedures to avoid complications. Preoperative imaging and careful planning are essential in such cases. In liver surgeries (e.g., segmentectomy or lobectomy), the absence of hepatic vein branches in segments 5 and 6 can increase the risk of complications. Lack of venous drainage may lead to venous stasis (blood pooling) in these segments.
\end{itemize}

Although often asymptomatic, awareness of this condition is crucial during surgical procedures 
to avoid complications.

\subsubsection{Set 25: Post-Operative Observations}

\begin{figure}[t]
    \centering
    \subfloat[]{\includegraphics[width=0.377\columnwidth]{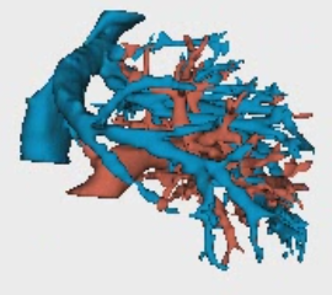}}
    \hfill
    \subfloat[]{\includegraphics[width=0.6\columnwidth]{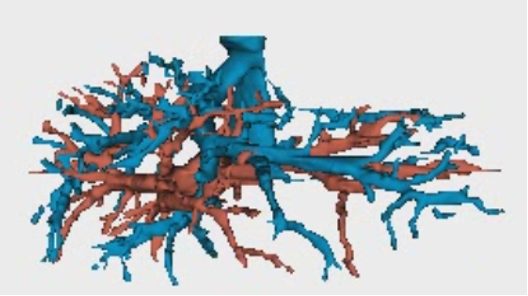}}
    
    \vspace{1em} 
    
    \subfloat[]{\includegraphics[width=0.497\columnwidth]{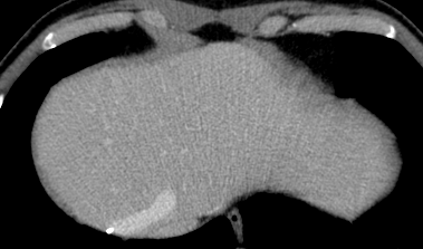}}
    \hfill
    \subfloat[]{\includegraphics[width=0.49\columnwidth]{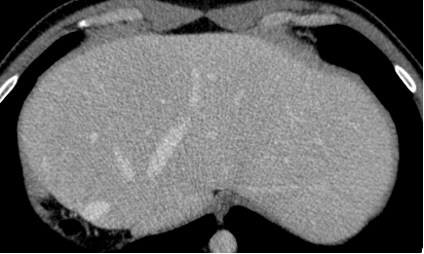}}

    \vspace{1em} 
    
    \subfloat[]{\includegraphics[width=0.4595\columnwidth]{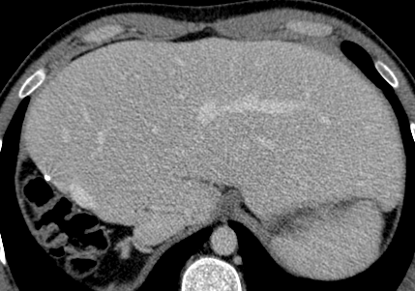}}
    \hfill
    \subfloat[]{\includegraphics[width=0.53\columnwidth]{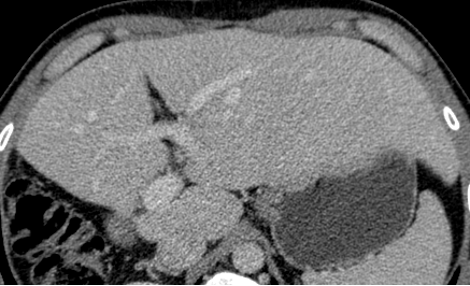}}

    \caption{Set 25. Right hepatectomy. The hepatic flexure occupies the space of the resected 
    right hepatic lobe, and the left lobe has expanded to compensate.}
    \label{fig:post_op_hepatectomy}
\end{figure}

In post-op CT imaging, certain characteristic findings are observed following a right hepatectomy (surgical removal of the right liver lobe) (Fig. \ref{fig:post_op_hepatectomy}). These findings help assess normal postoperative changes and detect potential complications. Below are the key observations on post-op CT after a right hepatectomy: the right lobe is surgically removed, leaving an empty space that is filled by the ascending colon and hepatic flexure. The left liver lobe often undergoes compensatory hypertrophy to adapt to the loss of the right lobe. The left lobe and caudate lobe are expected to grow over time, increasing their functional capacity to compensate for the resected lobe. The absence of the right hepatic vein and right portal vein can be observed. The remaining branches of the left hepatic vein and portal vein may show compensatory enlargement. In post-right hepatectomy CT, the remaining liver lobes, the surgical site, and vascular anatomy should be carefully evaluated for hypertrophy. Early detection of complications, such as bile leaks, vascular thrombosis, or abscesses, is crucial.

\subsubsection{Set 28: Portal Vein Branching Abnormality}

\begin{figure}[t]
    \centering
    \subfloat[]{\includegraphics[width=0.4664\columnwidth]{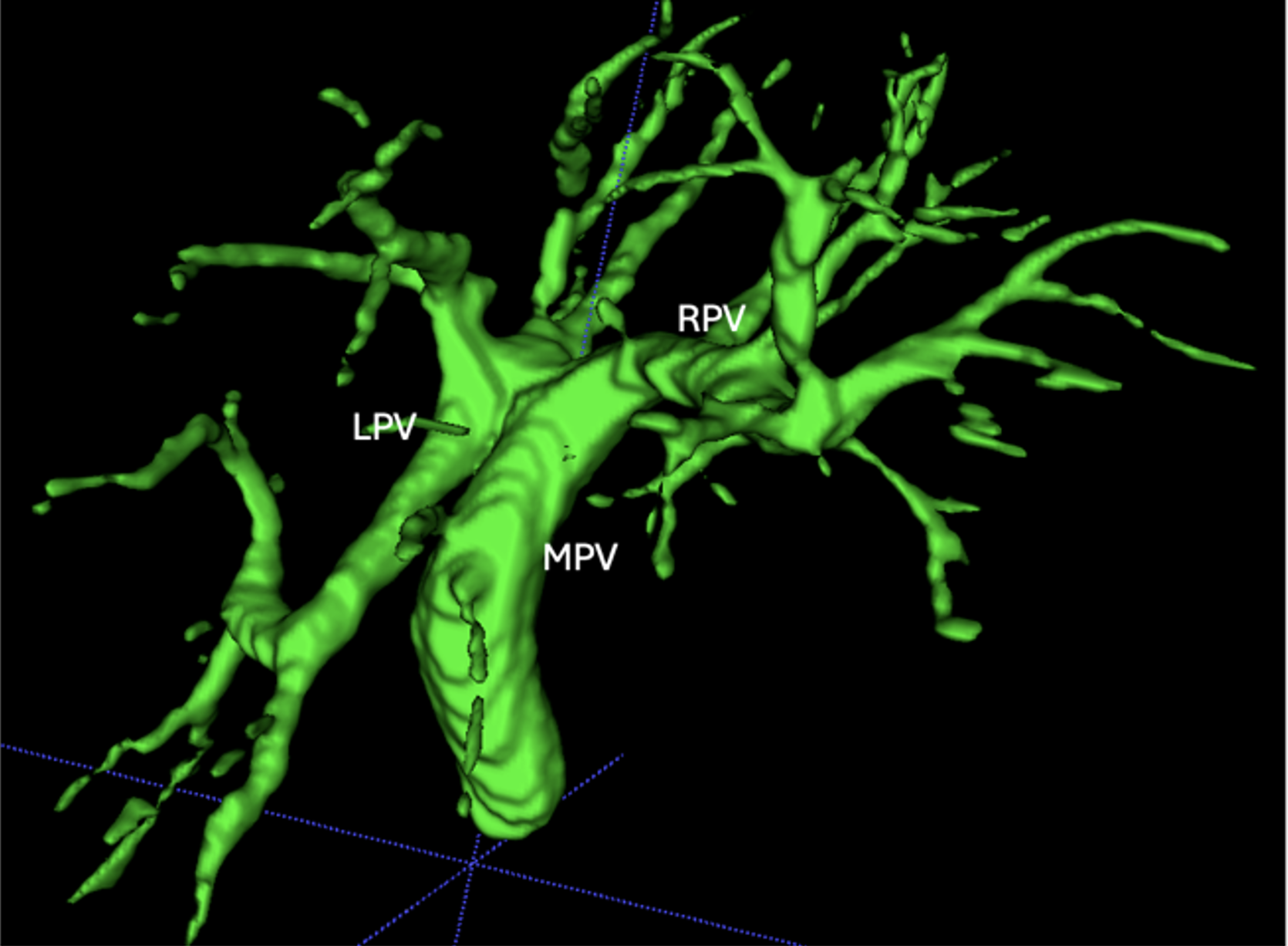}}
    \hfill
    \subfloat[]{\includegraphics[width=0.5163\columnwidth]{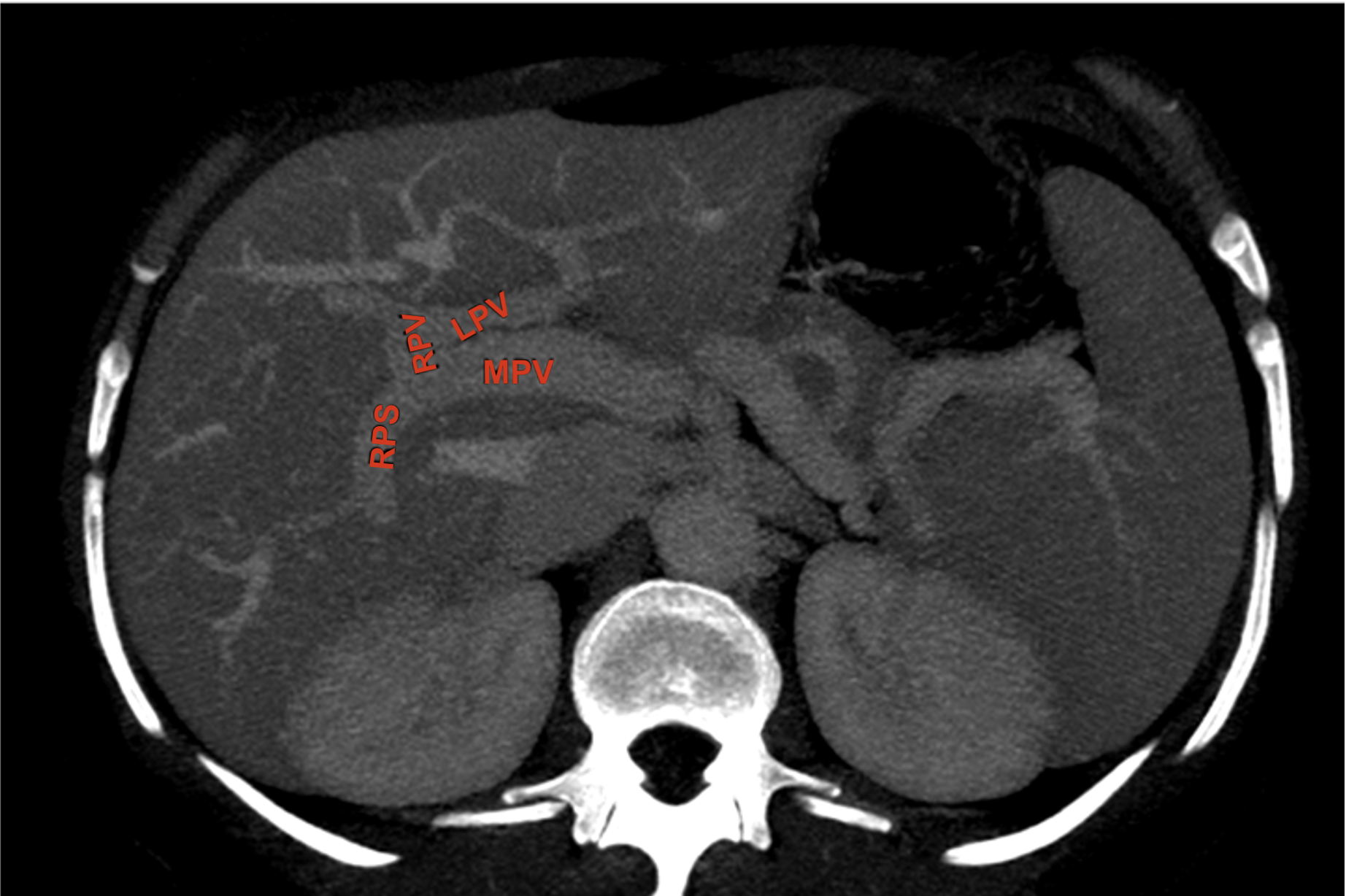}}
    \caption{Set 28. Portal vein branching anomaly: The LPV arises from the RPV. The MPV continues 
    as the RPV, which divides into right posterior (RPS) and right anterior (RAS) sector branches. 
    The LPV then turns sharply from right to left.}
    \label{fig:portal_vein_branching}
\end{figure}

This refers to an anatomical variation of the portal vein (Fig. \ref{fig:portal_vein_branching}). In this variation, the left portal vein does not arise directly from the main portal vein trunk but instead branches off from the right portal vein. In such cases, blood flow to the left lobe of the liver is supplied through a branch of the right portal vein. 

This is a developmental variation resulting from anomalies during embryological development of the portal venous system. It is usually asymptomatic and discovered incidentally. Awareness of such variations is critical in liver surgeries (e.g., lobectomy or liver transplantation). Accidental ligation of a branch could lead to insufficient blood supply to the left lobe. 

Thus, such anatomical variations should be accurately identified before any procedure, and surgical planning should be adapted accordingly to avoid complications.

\section{Evaluation Methods and Metrics}
\label{evaluationmetric}

To comprehensively assess the performance of vessel segmentation models on the VEELA dataset, we adopt a multi-metric evaluation strategy. Specifically, IoU, clDice, and NSD metrics are employed, following the recommendations of the Metrics Reloaded \citep{maier2024metrics} framework, which advocates for task-specific, holistic evaluation. These metrics collectively capture different aspects of segmentation quality, including volumetric overlap, centerline accuracy, and boundary precision. In addition to these established metrics, we discuss using Area and Length measures to further characterize segmentation outputs. 

Combining multiple evaluation metrics is important, especially in clinical applications where the segmentation requirements are multifaceted. For example, a model achieving high volumetric overlap (IoU) might still fail to accurately capture thin, elongated vessel structures critical for surgical planning, an issue better revealed by clDice and Length measures. Similarly, NSD can highlight subtle but clinically significant deviations along vessel boundaries that would otherwise go unnoticed in traditional overlap-based evaluations. We believe that by leveraging a diverse set of metrics, we can ensure a more robust, clinically meaningful assessment of model performance, ultimately facilitating the development of segmentation tools that are not only accurate but also reliable and trustworthy for clinical decision-making.

In the following section, we provide a detailed description of clDice, IoU, and NSD. We explain why integrating these metrics is particularly advantageous for clinical applications. Additionally, we compare the Area and Length measures against these metrics, including illustrative examples.

\subsection{Centerline Dice Similarity Coefficient (clDice)}
\label{subsec:cldice}

\begin{figure}[t]
    \centering
    \subfloat[]{\includegraphics[width=0.2007\columnwidth]{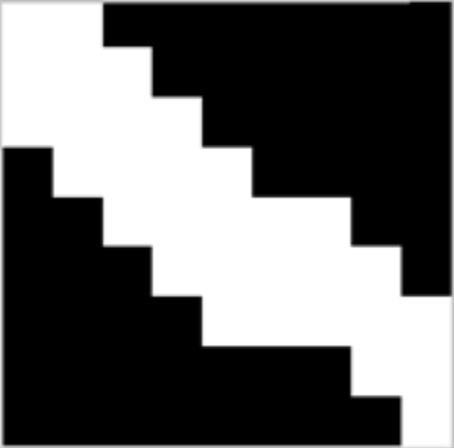}}
    \hfill
    \subfloat[]{\includegraphics[width=0.2005\columnwidth]{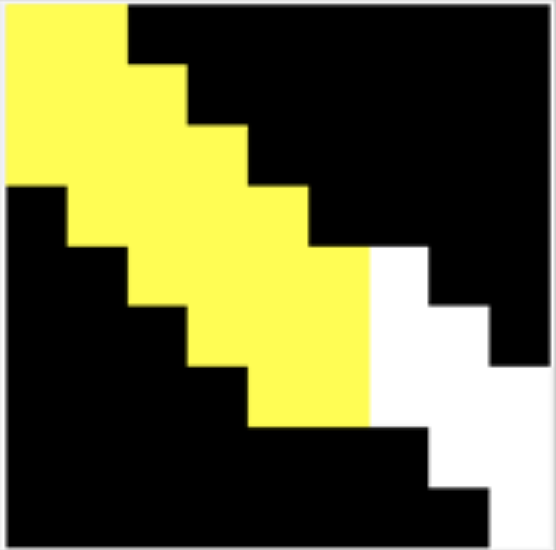}}
     \hfill
    \subfloat[]{\includegraphics[width=0.2\columnwidth]{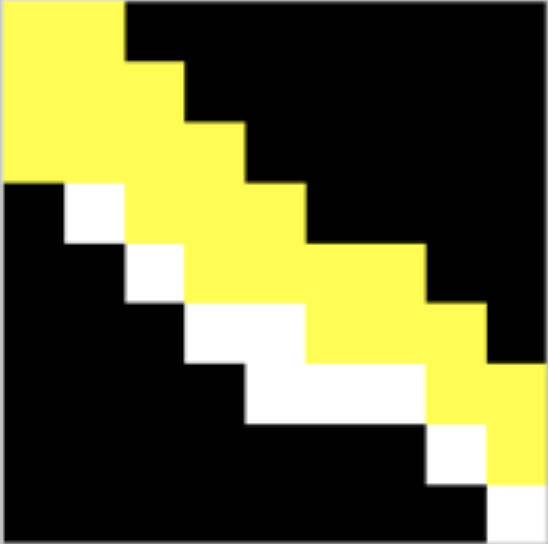}}
    \hfill
    \subfloat[]{\includegraphics[width=0.2005\columnwidth]{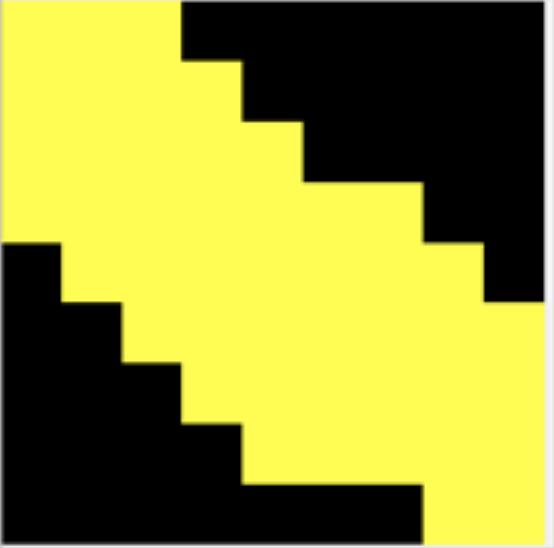}}
     \hfill
    \subfloat[]{\includegraphics[width=0.198\columnwidth]{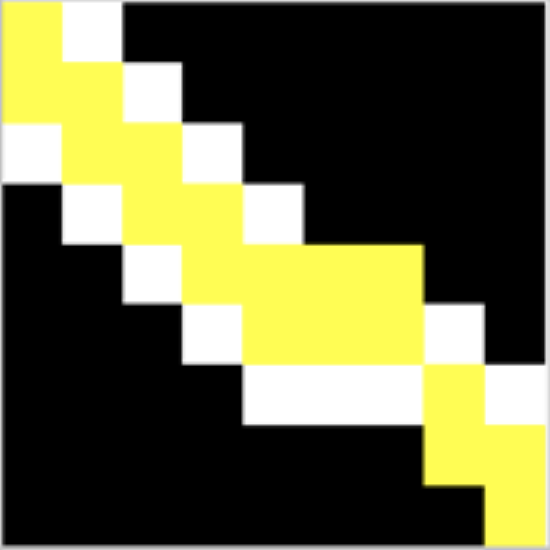}}
   
    \caption{Comparison of vessel segmentation performance with reference segmentation and corresponding Dice Similarity Coefficient (DSC) values. (a) Reference segmentation. (b) Segmentation result (yellow) with DSC = 0.8302, not preferred by clinicians despite high DSC. (c) Under-segmentation result (yellow) with DSC = 0.8302, preferred by clinicians due to clinically accurate delineation. (d) Over-segmentation result (yellow) with DSC = 0.8052, also preferred by clinicians for clinical relevance. (e) Under-segmentation result (yellow) with DSC = 0.7347, preferred by clinicians despite lower DSC.
    This highlights that while DSC is commonly used by researchers to evaluate segmentation performance, it is not always a suitable metric for capturing clinical preferences in vessel segmentation.}
    \label{fig:dsc_comp}
\end{figure}

The centerline Dice (clDice) metric was developed specifically to evaluate the topological accuracy of vessel segmentations by focusing on the alignment of the vessel skeletons. Let \(S_{\text{pred}}\) and \(S_{\text{ref}}\) denote the skeletons of the predicted segmentation and the ground truth, respectively. Two key components are defined as:
\begin{equation}
    T_{\text{prec}}(S_{\text{pred}}, S_{\text{ref}}) = \frac{|S_{\text{pred}} \cap S_{\text{ref}}|}{|S_{\text{pred}}|},
\end{equation}
\begin{equation}
    T_{\text{sens}}(S_{\text{ref}}, S_{\text{pred}}) = \frac{|S_{\text{ref}} \cap S_{\text{pred}}|}{|S_{\text{ref}}|}.
\end{equation}
The clDice is then computed as:
\begin{equation}
    \text{clDice}(S_{\text{ref}}, S_{\text{pred}}) = \frac{2 \times T_{\text{sens}}(S_{\text{ref}}, S_{\text{pred}}) \times T_{\text{prec}}(S_{\text{pred}}, S_{\text{ref}})}{T_{\text{sens}}(S_{\text{ref}}, S_{\text{pred}}) + T_{\text{prec}}(S_{\text{pred}}, S_{\text{ref}})}.
\end{equation}
clDice is highly effective at ensuring that the essential vessel network is captured, as it focuses on connectivity and branching structures. This topological focus addresses a critical limitation of traditional overlap-based metrics, such as the Dice Similarity Coefficient (DSC). As illustrated in Fig. \ref{fig:dsc_comp}, a high DSC score does not always correlate with clinical preference; for instance, a segmentation may achieve a high DSC while failing to capture essential connectivity or including clinically irrelevant artifacts (Fig. \ref{fig:dsc_comp}b). 

Conversely, clinicians often prefer results with lower DSC values if they maintain structural integrity and accurate delineation of the vascular tree (Fig. \ref{fig:dsc_comp}c--\ref{fig:dsc_comp}e). By prioritizing the skeleton, clDice provides a more robust assessment of the vessel network's continuity, which is often more valuable for clinical decision-making than pure voxel-wise overlap. However, its exclusive reliance on the skeleton means that it does not penalize errors in vessel thickness or minor positional shifts of the boundaries.

\subsection{Intersection over Union (IoU)}
\label{subsec:iou}

The Intersection over Union (IoU), also known as the Jaccard Index, is a widely recognized metric for measuring segmentation overlap. For any two sets \(A\) and \(B\) (which can be binary masks or bounding boxes), IoU is defined as:
\begin{equation}
    \text{IoU}(A, B) = \frac{|A \cap B|}{|A \cup B|},
\end{equation}
where \(|\cdot|\) denotes the cardinality (or area/volume). IoU offers a straightforward quantitative measure of overall segmentation accuracy; however, it does not capture shape nuances or minor boundary discrepancies that are often critical for segmenting thin and branching vascular structures.

\subsection{Normalized Surface Distance (NSD)}
\label{subsec:nsd}

The Normalized Surface Distance (NSD) metric, also referred to as Surface Dice, evaluates the precision of segmentation boundaries by measuring the proportion of boundary pixels (or voxels) that lie within a specified tolerance \(\tau\) of the corresponding ground truth boundaries. Let \(S_A\) and \(S_B\) denote the sets of boundary pixels of the predicted segmentation and the ground truth, respectively, and let \(S_A^{(\tau)}\) denote the dilated boundary of \(S_A\) by \(\tau\). NSD is then defined as:
\begin{equation}
    \text{NSD}(A, B) = \frac{|S_A^{(\tau)} \cap S_B| + |S_B^{(\tau)} \cap S_A|}{|S_A| + |S_B|}.
\end{equation}
This metric allows us to quantify how closely the predicted boundaries align with the ground truth within a clinically acceptable margin of error. The choice of \(\tau\) is critical; too large a value may mask significant segmentation errors, while too small a value may penalize minor deviations that are not clinically relevant.

\subsection{Area and Length Measures}
\label{subsec:area_length}

\begin{figure}[t]
    \centering
    \subfloat[\label{fig:mfig1a}]{\includegraphics[width=\columnwidth]{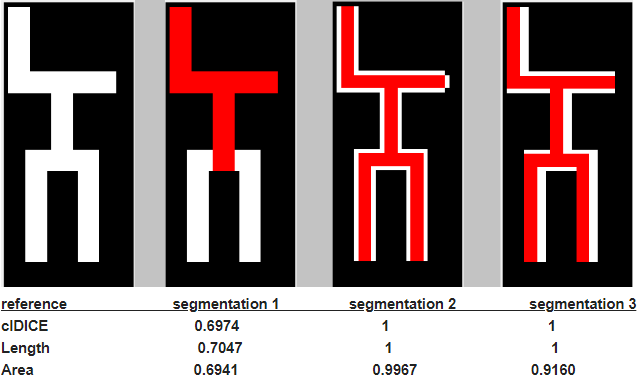}}
    \vspace{1em} 
    \subfloat[\label{fig:mfig1b}]{\includegraphics[width=\columnwidth]{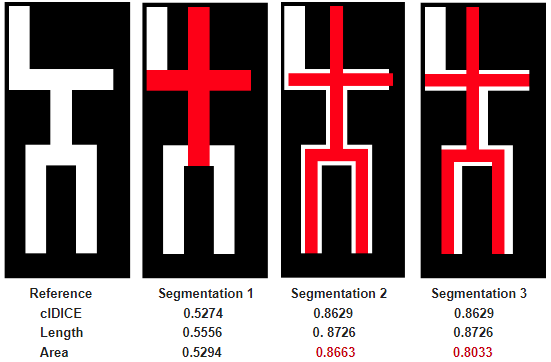}}
       \caption{A representative figure of comparison between area, length, and clDice metrics.}
        \label{fig:metric1}
    \end{figure}

The Area and Length measures (Fig. \ref{fig:metric1}) are designed to assess two essential aspects of vessel segmentation: the faithful reproduction of vessel boundaries (and hence thickness) and the complete capture of the vessel's longitudinal extent. These measures provide critical information that complements topology‐based metrics.

\paragraph{Area Measure}  
The Area measure quantifies the overlap between the predicted segmentation \(S\) and the ground truth \(S_G\) by comparing the 2D cross-sectional areas of the vessels. To account for minor discrepancies in boundary delineation, a morphological dilation operator \(\delta_\alpha\) (with dilation radius \(\alpha\)) is applied to both \(S\) and \(S_G\). This yields a tolerance for slight variations in vessel thickness that may arise from differences in manual annotation. Formally, the Area measure is defined as:
\begin{equation}
    \text{Area} = \frac{\#\Big(\big(\delta_\alpha(S) \cap S_G\big) \cup \big(S \cap \delta_\alpha(S_G)\big)\Big)}{\#\big(S \cup S_G\big)},
\end{equation}
where \(\#(\cdot)\) denotes the number of pixels (or voxels). A properly tuned \(\alpha\) allows the metric to indicate whether the vessel's thickness is over- or under-segmented, a detail of high clinical importance when assessing lumen sizes for interventions or surgical planning.

\paragraph{Length Measure}  
The Length measure assesses whether the segmentation captures the vessel's full longitudinal course. Given that small lateral shifts may occur without compromising the overall vessel trajectory, we first extract the skeleton (or centerline) \(\sigma(\cdot)\) of the segmented vessel and then apply a tolerance via a dilation operator \(\delta_\beta\) (with radius \(\beta\)). The Length measure is expressed as:
\begin{equation}
    \text{Length} = \frac{\#\Big(\big(\sigma(S) \cap \delta_\beta(S_G)\big) \cup \big(\delta_\beta(S) \cap \sigma(S_G)\big)\Big)}{\#\big(\sigma(S) \cup \sigma(S_G)\big)}.
\end{equation}
This metric ensures that the predicted segmentation not only preserves connectivity but also accurately represents the vessel’s full spatial extent, which is critical in applications such as determining resection margins or guiding interventional tools.

\paragraph{Rationale: Area and Length versus clDice}  
While the clDice metric is adept at ensuring that the connectivity of the vessel network is maintained, it inherently focuses on the alignment of the vessel skeletons. This means that clDice may report high scores even when the segmented vessel boundaries are significantly misaligned in terms of thickness. For example, a segmentation that is either too thin or slightly shifted laterally may still yield a perfect clDice score if the skeletons overlap completely. However, such discrepancies are clinically significant; an under‐segmented vessel might imply an erroneously narrow lumen, and an over‐segmented vessel might obscure adjacent anatomical structures. Therefore, Area and Length measures are indispensable in providing additional sensitivity to the actual vessel width and full extent, ensuring that the segmentation is not only topologically correct but also geometrically faithful.

\subsection{Summary and Clinical Relevance}
\label{subsec:summary}

In summary, our comprehensive evaluation framework combines multiple metrics to capture both the topological and geometrical fidelity of vessel segmentations:
\begin{itemize}
    \item \textbf{clDice} excels at ensuring that the overall vessel connectivity and branching structure are intact, making it highly valuable for verifying that no critical vessel segments are missing.
    \item \textbf{IoU} offers a straightforward measure of overall segmentation accuracy, while \textbf{NSD} provides a refined assessment of boundary precision.
    \item \textbf{Area Measure} directly assesses the accuracy of vessel thickness by evaluating the overlap of vessel surfaces. This is essential for determining lumen size and ensuring that the segmentation reflects the true vessel boundaries.
    \item \textbf{Length Measure} verifies that the full longitudinal extent of the vessel is captured, ensuring that distal branches are not omitted and that the segmentation follows the vessel path accurately.
    
\end{itemize}

From a clinical perspective, an automated vessel segmentation method must not only produce a connected vessel network, but also accurately represent the physical dimensions and full extent of the vessels. For instance, an under‐segmented vessel might misleadingly suggest a narrowed lumen, while an over‐segmented vessel might obscure adjacent tissue boundaries. Thus, while clDice is indispensable for checking connectivity, the Area and Length measures provide critical complementary insights that together yield a more complete and clinically meaningful evaluation of segmentation quality.

This multifaceted evaluation strategy ensures that our segmentation methods meet the stringent requirements of clinical applications, ultimately improving preoperative planning, surgical navigation, and patient outcomes.

\section{Methodologies}
\label{propsedmet}


To promote algorithmic innovation, the VEELA dataset was organized as a grand challenge, offering a standardized platform for researchers to benchmark their automatic vessel segmentation and classification methods. Participants were not limited to using only the provided VEELA 2025 training data, but were allowed to use pre-trained models or supplemental datasets. 

The challenge was organized into two tasks: Segmentation (Task 1) and Classification (Task 2). In Task 1, participants were required to provide model predictions as a single binary mask identifying the combined regions of hepatic and portal vessels. In Task 2, the goal was to classify the binary mask to differentiate between the hepatic and portal vessels.  

The results of all teams have been published on Synapse. In the following sections, we describe the architectures and optimization methods used by the leading teams: UW-Madison-AIR, Haqan, and MedInsight-ViseurAI. We will also examine the hybrid volumetric models, including Swin UNETR \citep{hatamizadeh2021swin}, GLIMS \citep{yazici2024glims}, and VSNet \citep{xu2025vsnet}, which were employed to assess baseline performance.

\subsection{UW-Madison-AIR}

Team UW-Madison-AIR~\citep{seker2025hepatic} developed their model based on the nnU-Net \citep{isensee2021nnu} framework, which was specifically enhanced for the VEELA 2025 challenge. The training process incorporated a combination of CTA and contrast-enhanced CT images obtained from the Medical Segmentation Decathlon, LIRCAD, and VEELA 2025 datasets. A total of five nnU-Net-based models were tested, each distinguished by its loss function. These designs included variations of DSC, cross-entropy loss, clDice to promote topological consistency, and edge map Dice loss to improve boundary delineation. The final selected model used a weighted composite loss consisting of 70\% cross-entropy loss, 15\% Dice segmentation loss, and 15\% clDice. This weighting prioritized voxel-wise hepatic-versus-portal vessel classification while also preserving regional overlap and vessel topology. One of the key methodological contributions was the introduction of a centerline-regression cross-entropy loss to improve the classification of small-caliber vessels. Additionally, clDice was computed using a 3D thinning algorithm to accurately extract vessel centerlines.

\subsection{Haqan}

Team Haqan used the nnU-Net framework's full-resolution 3D configuration to automate segmentation of hepatic and portal veins in contrast-enhanced CTA images. The model was trained for over 250 epochs on a combined dataset comprising the Medical Segmentation Decathlon~\citep{antonelli2022medical} (Task 08 – Hepatic Vessels) and the VEELA 2025 training set. Preprocessing included resampling all images to a target voxel spacing of approximately [1.5, 0.8, 0.8] mm and applying CT-specific intensity normalization. The training process incorporated extensive data augmentation, including random rotations, scaling, elastic deformations, gamma correction, and mirroring, supplemented with contrast and brightness adjustments to simulate variability in CTA image characteristics. For post-processing, connected component analysis was applied, retaining only the largest component per class to suppress false positives. No external datasets or pretrained models were used in this approach.

\subsection{MedInsight-ViseurAI}

Team MedInsight-ViseurAI proposed a 3D U-Net \citep{cciccek20163d} model with residual connections for vessel segmentation in the VEELA dataset. CT volumes were normalized using liver masks, and 64×128×128 patches were extracted via a sliding window with 32×64×64 strides. The model follows an encoder-decoder architecture with four levels, using residual convolutional blocks, skip connections, and a 1024-channel bottleneck layer. Model training utilized a Boundary Tversky Loss, which integrates Tversky loss ($\alpha$=0.7, $\beta$=0.3) to address class imbalance, and a boundary-sensitive Difference of Union (DoU) term ($\lambda$=0.5, $\gamma$=0.8) to improve segmentation precision near vessel edges. Training was performed for up to 300 epochs using the Adam optimizer, with a learning rate reduction on a learning rate plateau. Validation was performed after each epoch, and training stopped when the validation loss dropped below 0.01.

\subsection{Baseline}
\label{propsedmetbaseline}

As baseline approaches, we benchmark the VEELA dataset using three hybrid volumetric semantic segmentation models. This selection includes Swin UNETR \citep{hatamizadeh2021swin} and GLIMS \citep{yazici2024glims, yazici2023attention}, which have demonstrated robust performance in general volumetric organ segmentation, alongside the VSNet \citep{xu2025vsnet}, which is specifically engineered to maintain the topological connectivity of complex vascular structures. Together, these models provide a comprehensive evaluation of how general-purpose and structure-aware architectures perform on the intricate vessel geometries of the VEELA dataset.

\subsubsection{Deep Learning Architectures}

Swin UNETR is a transformer-based model for 3D semantic segmentation. It features a U-shaped architecture with a Swin Transformer encoder and a CNN-based decoder, utilizing skip connections for multi-resolution processing. Leveraging a shifted windowing mechanism, it captures long-range dependencies and multi-scale features. It is widely used in state-of-the-art architectures for its efficiency and robustness. However, it is insufficient for fine-grained vessel segmentation.

Similar to Swin UNETR, GLIMS also utilizes a hybrid convolutional neural network and transformer-based model designed for efficient and accurate 3D medical image segmentation. It integrates dilated convolutional blocks (DACB) to capture local and global features and employs Swin Transformer \citep{liu2021swin} bottlenecks to model long-range dependencies, thereby enhancing the fusion of detailed spatial and contextual information. GLIMS also introduces Channel- and Spatial-Wise Attention Blocks (CSAB) to guide the decoder to focus on relevant regions during segmentation. Despite its lightweight architecture, with significantly fewer parameters and lower computational cost than state-of-the-art models, GLIMS achieves superior performance on glioblastoma and multi-organ segmentation datasets, making it especially effective in scenarios with limited training data.

VSNet is a multi-task learning framework for segmenting the complex structures of hepatic and portal veins, particularly minor vessels. Its key innovation is a vessel-growing decoder that simulates the natural extension of minor vessels from main trunks, ensuring topological connectivity and preventing misclassification. This is supported by a Global Dual Transformer (GDT) to capture spatial dependencies and a Dependency Transmitting (DT) block that transmits these relationships to upsampled features, thereby guiding secondary vessel growth. Additionally, VSNet employs dense centerline regression and edge segmentation as auxiliary tasks to maintain the integrity and boundaries of vascular structures. By integrating deep supervision with these components, VSNet preserves the liver's complex vascular anatomy.

\subsubsection{Augmentation and Post-Processing Techniques}

To enhance the segmentation performance of CT scan models, we apply multiple preprocessing and postprocessing techniques, such as anisotropic sampling and liver masking, to both the input and output scans. These include data augmentation to improve data distribution and fine-grained error correction for mask predictions.

Anisotropic pre-processing of CT scans involves resampling images with non-uniform voxel dimensions to achieve consistent spatial resolution across all axes. This step is crucial in medical imaging analysis, particularly in deep learning applications, to ensure that models accurately interpret anatomical structures without distortion due to varying voxel sizes. To benefit from anisotropic sampling, we automatically select resampling strategies based on the degree of anisotropy, as described in \citep{isensee2021nnu}.

\begin{figure}[t!]
    \centering
    \subfloat[]{\includegraphics[width=0.5\columnwidth]{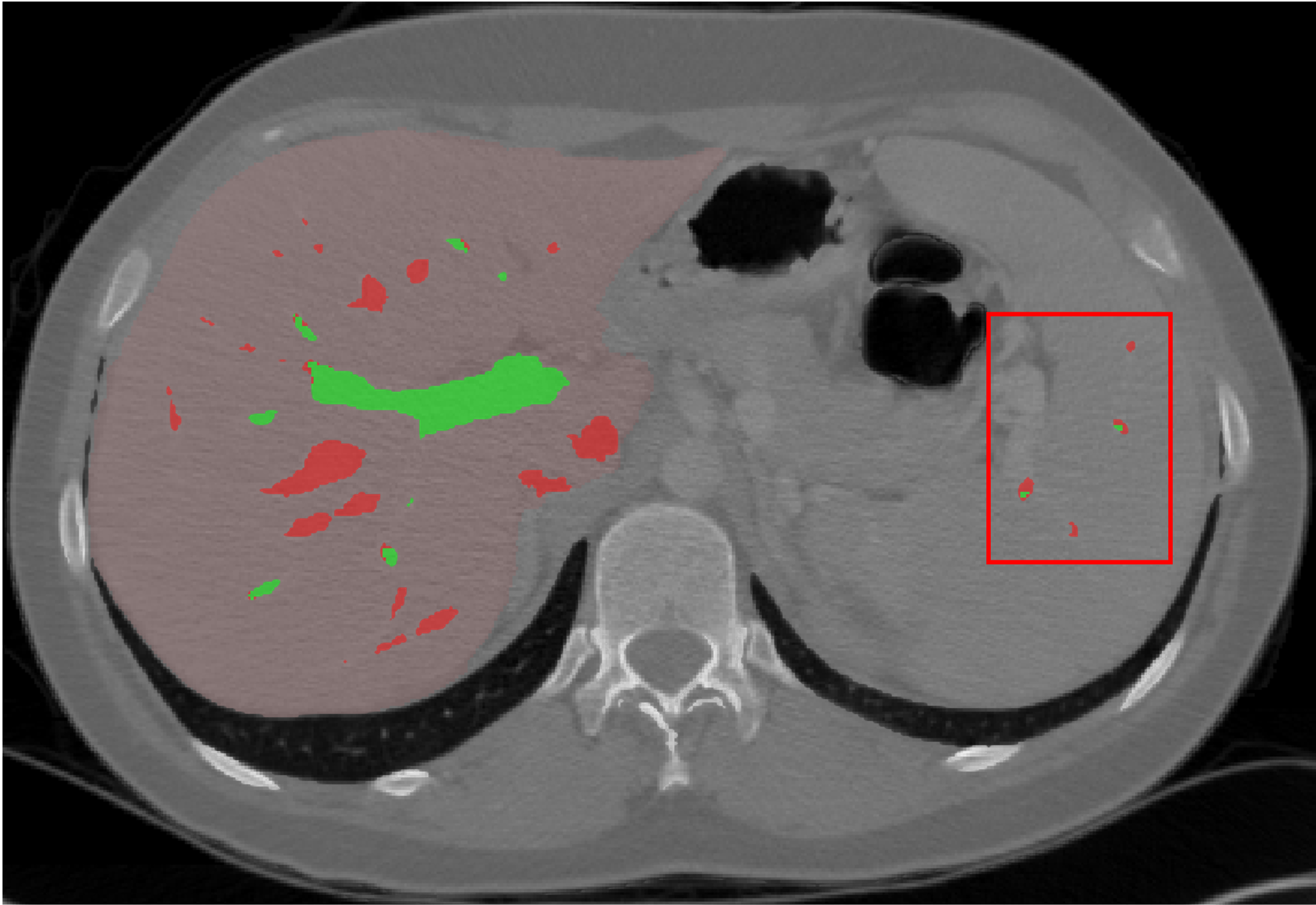}}
    \hfill
    \subfloat[]{\includegraphics[width=0.5\columnwidth]{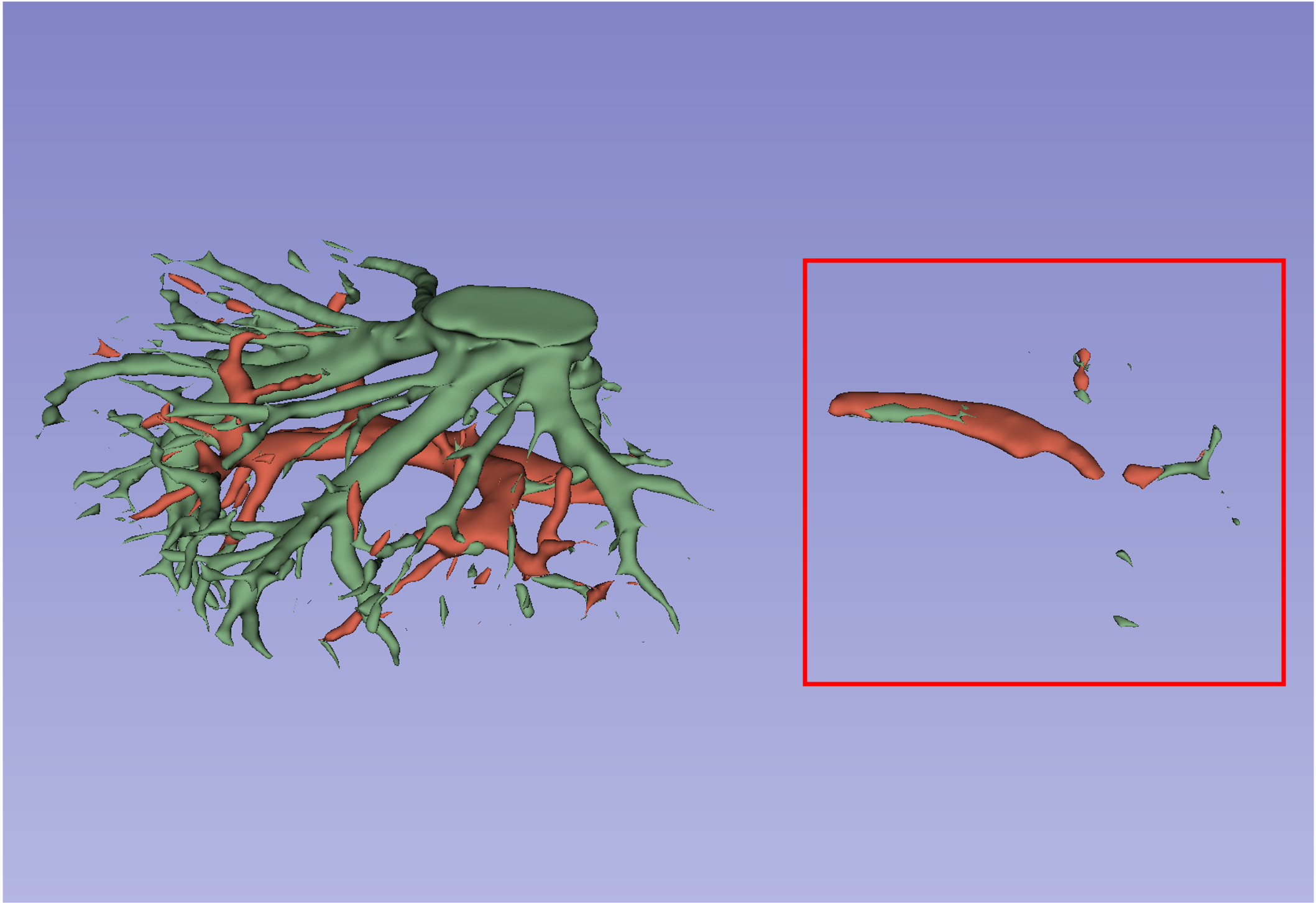}}
     \caption{The predicted vessels both on and outside the liver region (a). The predictions for vessels in the red bounding boxes have been removed using AND gating with the predicted liver mask, which is overlaid in light red. The 3D visualization of the correct and mispredicted vessels (b).}
    \label{fig:postprocessing_liver_mask}
\end{figure}

When predicting existing hepatic and portal vessel structures in CT scans, regions outside of the liver may also be identified as potential vessel structures due to similarities in intensities and shapes (Fig. \ref{fig:postprocessing_liver_mask}). To improve vessel segmentation, we perform a liver segmentation task, using the output mask alongside the vessel predictions via AND gating. As a result, we could reduce the number of false-positive predictions, thereby improving the clDice, IoU, and NSD metrics.

\section{Results}
\label{applications}

In this section, we present the results of VEELA 2025, together with the team submissions and the baseline models.

The evaluations of team submissions were conducted on the challenge server via Synapse, using 20 CTA cases across two tasks: Segmentation (Task 1) and Classification (Task 2). Participants submitted multiple entries, and the best result from each participant was used to determine the final ranking. The final rankings were calculated using the arithmetic mean of the clDice, IoU, and NSD metrics. In total, there were 17 registrations across 4 teams that made valid submissions\footnote{Participating teams: \url{https://www.synapse.org/Synapse:syn65471967/wiki/632178}.}.

\begin{table}[!t]
\centering
\caption{Averaged vessel segmentation metrics for the baseline models on the test set. Best results are in bold, and second-best are underlined.}
\label{tab:results}
\begin{tabular}{@{}lcccc@{}}
\hline
\textbf{Method} & \textbf{Liver Mask} & \textbf{clDice} & \textbf{IoU} & \textbf{NSD} \\ \hline
Swin UNETR      & \texttimes & 0.688 & 0.525 & 0.653 \\
                & \checkmark & \underline{0.695} & \underline{0.530} & 0.660 \\
GLIMS           & \texttimes & 0.756 & 0.580 & 0.727 \\
                & \checkmark & \textbf{0.769} & \textbf{0.588} & \textbf{0.734} \\
VSNet           & \texttimes & 0.649 & 0.486 & 0.637 \\
                & \checkmark & 0.662 & 0.496 & \underline{0.668} \\ \hline
\end{tabular}
\end{table}

\subsection{Baseline Results}
\begin{figure}[!t]
    \centering
    \includegraphics[width=\columnwidth]{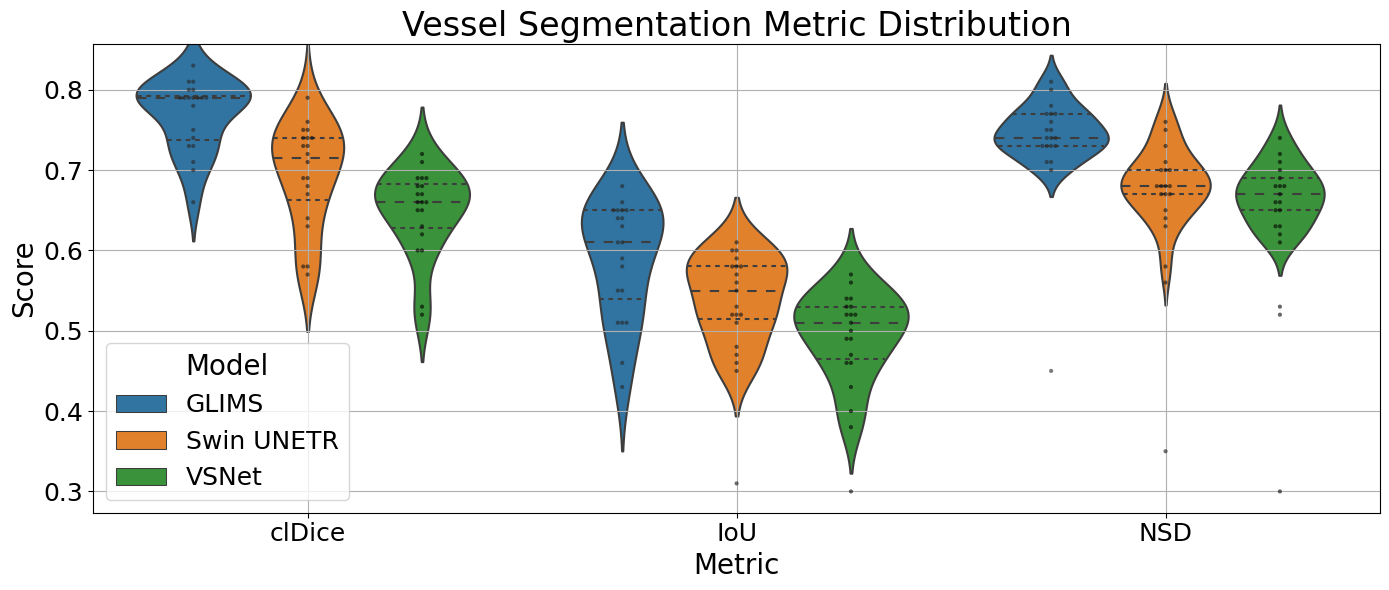}
     \caption{The metric results for the test split that includes 20 samples. It was observed that the GLIMS model outperformed Swin UNETR and VSNet in terms of average clDice, IoU, and NSD. The plots indicate that the Swin UNETR and VSNet samples exhibit greater deviation from the mean, whereas GLIMS shows more consistent performance.}
    \label{fig:violin_results}
\end{figure}

\begin{figure}[!t]
    \centering
    \includegraphics[width=\columnwidth]{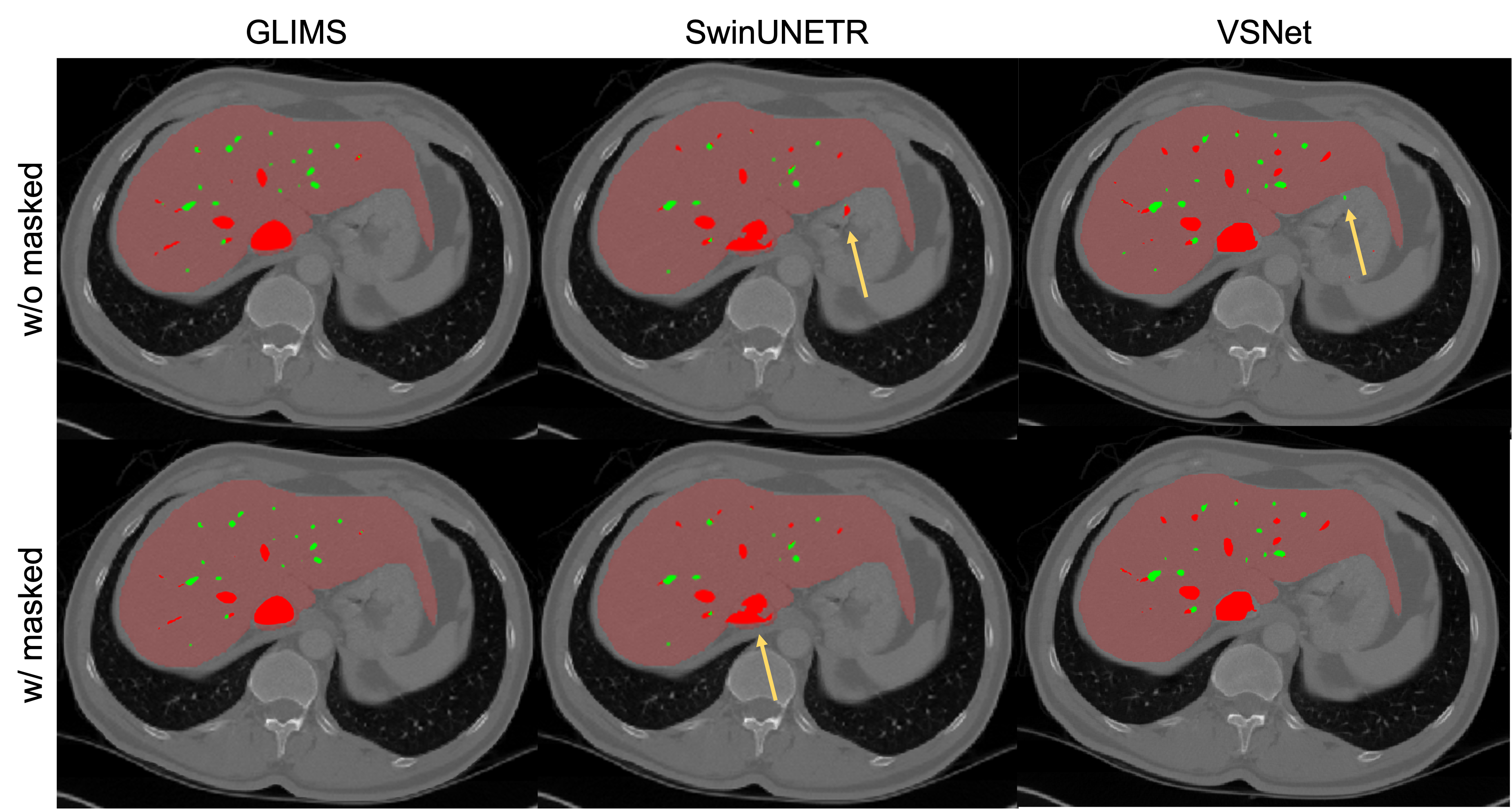}
     \caption{The qualitative results for test sample 17 using the GLIMS, Swin UNETR, and VSNet models, with and without liver-masked post-processing. As indicated by the yellow arrows, the Swin UNETR and VSNet models tend to generate false-positive vessel masks outside the liver region and often produce incomplete masks in true-positive areas. In contrast, the GLIMS model demonstrates better overall performance.}
    \label{fig:Set17_samples}
\end{figure}

As mentioned in Section \ref{propsedmetbaseline}, we use the Swin UNETR, GLIMS, and VSNet models to present the benchmark results. The models were trained using 5-fold cross-validation and tested with consistent pre-processing and post-processing settings. The 5-fold weights of the models were combined as an ensemble for the test split. The mask prediction results are reported in Table \ref{tab:results} and Fig. \ref{fig:violin_results} using clDice, IoU, and NSD metrics. The results indicate that GLIMS outperformed the Swin UNETR and VSNet models in vessel segmentation, achieving a clDice score of 76.9\% with liver masks and 75.6\% without them. In comparison, Swin UNETR achieved clDice scores of 69.6\% with liver masks and 68.8\% without liver masks, while VSNet achieved 66.2\% with liver masks and 64.9\% without. The performance of the baseline models is also evaluated qualitatively. As shown in Fig. \ref{fig:Set17_samples}, the GLIMS model demonstrates precise segmentation results, both with and without liver masks. In contrast, the Swin UNETR and VSNet models exhibit discrepancies in the generated masks and tend to produce inaccurate predictions for the regions outside the liver.

\begin{table*}[t]
\centering
\caption{Classification Task Results: Hepatic vs. Portal. Area and Length metrics were calculated using a 5-pixel dilation. Top results are in bold, and the second-best results are underlined. The baseline method includes Liver Masking post-processing.}
\label{tab:Task2}
\begin{tabular}{llccccc}
\toprule
\textbf{Teams} & \textbf{Vessel} & \textbf{clDice} & \textbf{IoU} & \textbf{NSD} & \textbf{Area} ($\delta_5$) & \textbf{Length} ($\delta_5$) \\ \midrule
\multirow{2}{*}{UW-Madison-AIR} & Hepatic & \textbf{0.830 ± 0.03} & \underline{0.512 ± 0.07} & \textbf{0.697 ± 0.07} & \underline{0.837 ± 0.09} & \textbf{0.914 ± 0.02} \\
                                & Portal  & \textbf{0.704 ± 0.06} & \underline{0.371 ± 0.04} & \underline{0.625 ± 0.07} & \underline{0.632 ± 0.05} & \textbf{0.780 ± 0.04} \\ \addlinespace
\multirow{2}{*}{Haqan}          & Hepatic & 0.682 ± 0.05 & 0.490 ± 0.08 & 0.595 ± 0.06 & 0.774 ± 0.10 & 0.756 ± 0.05 \\
                                & Portal  & 0.588 ± 0.05 & 0.350 ± 0.04 & 0.550 ± 0.05 & 0.565 ± 0.05 & 0.658 ± 0.06 \\ \addlinespace
\multirow{2}{*}{MedInsight-ViseurAI} & Hepatic & 0.518 ± 0.04 & 0.385 ± 0.01 & 0.509 ± 0.04 & 0.698 ± 0.07 & 0.672 ± 0.06 \\
                                     & Portal  & 0.446 ± 0.06 & 0.352 ± 0.02 & 0.472 ± 0.06 & 0.621 ± 0.05 & 0.585 ± 0.07 \\ \midrule
\multirow{2}{*}{Baseline (GLIMS)}    & Hepatic & \underline{0.678 ± 0.04} & \textbf{0.565 ± 0.09} & \underline{0.684 ± 0.07} & \textbf{0.875 ± 0.08} & \underline{0.828 ± 0.04} \\
                                     & Portal  & \underline{0.634 ± 0.05} & \textbf{0.511 ± 0.07} & \textbf{0.643 ± 0.06} & \textbf{0.833 ± 0.03} & \underline{0.750 ± 0.04} \\ \bottomrule
\end{tabular}
\end{table*}

\begin{table*}[]
\centering
\caption{Segmentation Task Results: Vessels vs. Background. The scores were retrieved from the online evaluation server. Area and Length metrics were calculated after the challenge, using a 5-pixel dilation. The baseline method includes Liver Masking as a post-processing step. The top result is highlighted in bold, and the second-best result is underlined.}
\label{tab:Task1}
\begin{tabular}{lccccc}
\hline
\textbf{Teams}      & \textbf{clDice}       & \textbf{IoU}          & \textbf{NSD}          & \textbf{Area ($\delta_5$)} & \textbf{Length ($\delta_5$)} \\ \hline
UW-Madison-AIR      & 0.691 ± 0.07          & 0.505 ± 0.03          & 0.626 ± 0.08          & 0.703 ± 0.06                      & 0.738 ± 0.05                        \\
Haqan               & 0.662 ± 0.06          & 0.445 ± 0.04          & 0.593 ± 0.07          & 0.680 ± 0.05                      & 0.697 ± 0.04                        \\
MedInsight-ViseurAI & {\underline{0.731 ± 0.06}}    & {\underline{0.534 ± 0.07}}    & {\underline{0.704 ± 0.07}}    & {\underline{0.869 ± 0.05}}                & {\underline{0.853 ± 0.06}}                  \\ \hline
Baseline (GLIMS)    & \textbf{0.769 ± 0.04} & \textbf{0.588 ± 0.07} & \textbf{0.734 ± 0.07} & \textbf{0.910 ± 0.06}             & \textbf{0.893 ± 0.03}               \\ \hline
\end{tabular}
\end{table*}

\begin{figure*}[t]
    \centering
    \includegraphics[width=\textwidth]{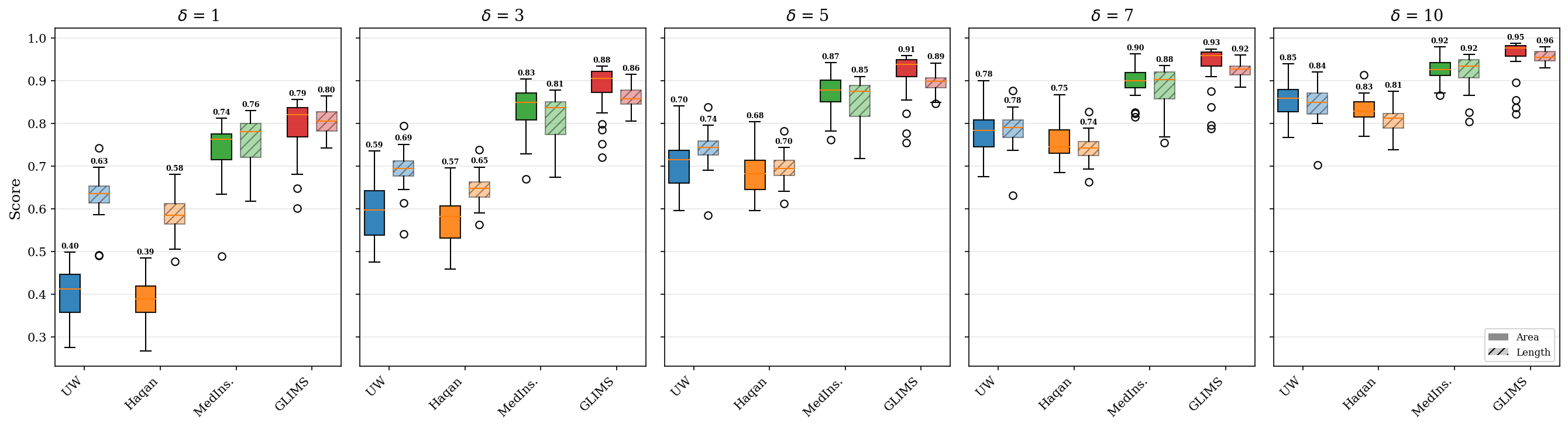}
     \caption{Distribution of Area and Length scores across dilation rates $\delta \in \{1, 3, 5, 7, 10\}$ for the Segmentation Task. Solid boxes represent Area, and hatched boxes represent Length. Mean values are annotated above each box. All teams exhibit monotonically increasing scores as $\delta$ increases, reflecting greater spatial tolerance in the matching criterion.}
    \label{fig:Area_Lenght_Dilation_Results}
\end{figure*}

\begin{figure}[t]
    \centering
    \includegraphics[width=\columnwidth]{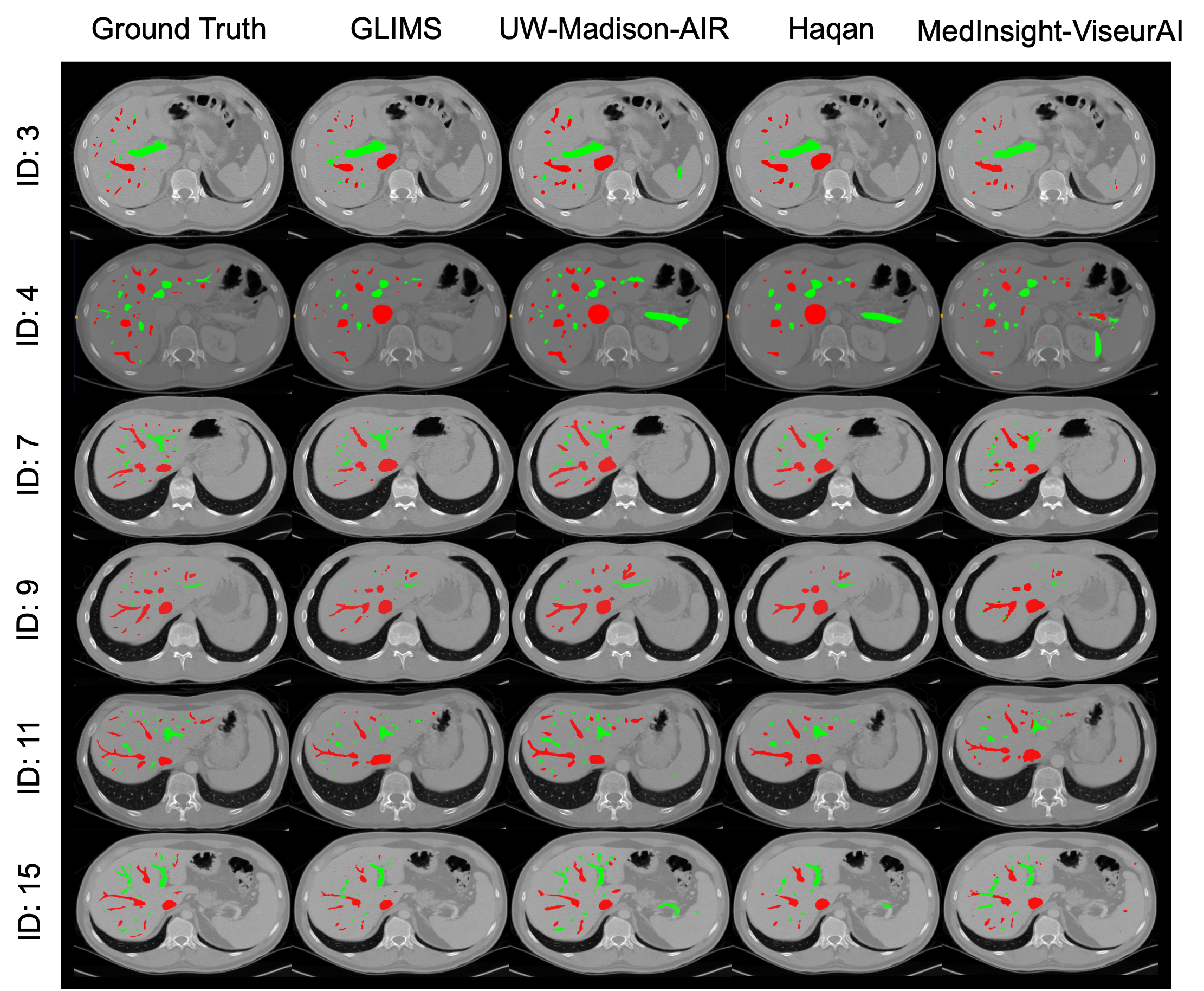}
     \caption{The qualitative results for the selected sample cases in the test set of VEELA 2025. The columns show the ground-truth mask and the predictions from GLIMS, UW-Madison-AIR, Haqan, and MedInsight-ViseurAI for the hepatic (red) and portal (green) vessels, respectively. The rows represent sample slices from the selected patients in the test set.}
    \label{fig:Qualitative_results}
\end{figure}

\begin{figure}[t]
    \centering
    \includegraphics[width=\columnwidth]{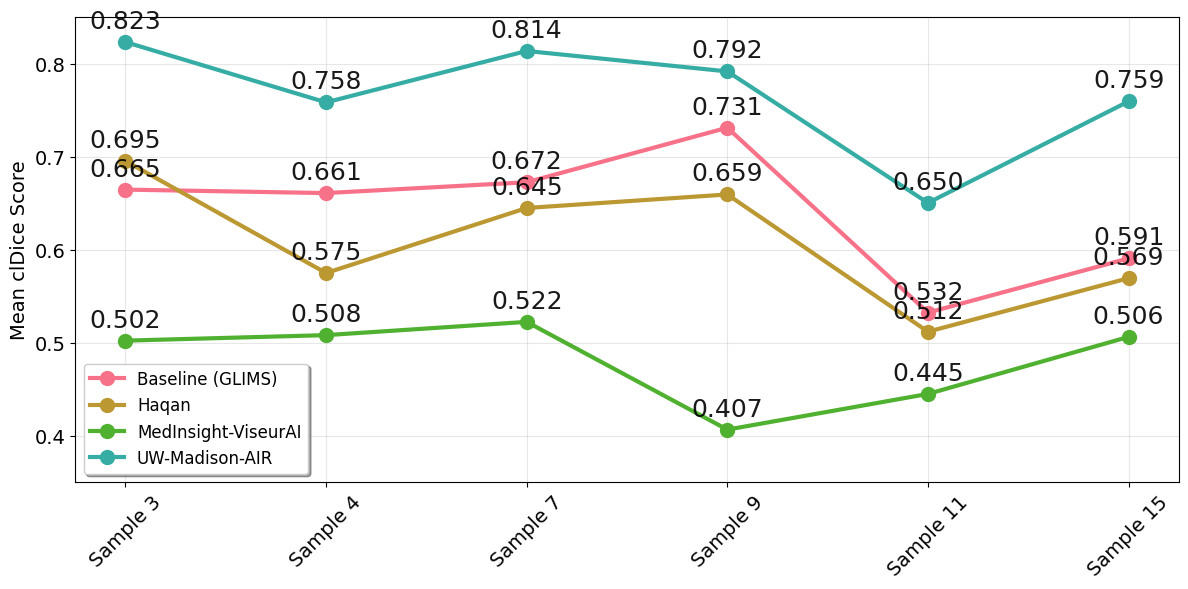}
     \caption{The mean clDice score results of the participants on the selected test set samples for the Segmentation Task.}
    \label{fig:Quantitative_results}
\end{figure}

\subsection{Challenge Results}
\label{challenge_results}
The test set scores for the challenge submissions are presented in Table~\ref{tab:Task2} for the Segmentation Task and in Table~\ref{tab:Task1} for the Classification Task. In the Segmentation Task, the MedInsight-ViseurAI team achieved the highest performance among participants, earning scores of 73.1\%, 53.4\%, and 70.4\% on the clDice, IoU, and NSD metrics, respectively. Notably, they also led the participant group in the geometric metrics, achieving 86.9\% in Area ($\delta_5$) and 85.3\% in Length ($\delta_5$). The second-place team, UW-Madison-AIR, followed with 69.1\%, 50.5\%, and 62.6\% for clDice, IoU, and NSD, respectively, and scored 70.3\% and 73.8\% on Area and Length ($\delta_5$) metrics.

The Area and Length metrics are evaluated across five dilation rates $\delta \in \{1, 3, 5, 7, 10\}$, where smaller values impose stricter spatial tolerance and larger values allow more lenient matching. As expected, all teams exhibit a monotonically increasing trend in both metrics as the dilation rate grows (Fig. \ref{fig:Area_Lenght_Dilation_Results}). For MedInsight-ViseurAI, the Area score rises from 73.6\% at $\delta_1$ to 92.5\% at $\delta_{10}$, while Length increases from 75.8\% to 91.9\% over the same range. Similarly, UW-Madison-AIR improves from 39.8\% to 85.3\% in Area and from 62.8\% to 84.2\% in Length across dilation rates. Notably, the performance gap between teams narrows considerably at larger dilation rates. At $\delta_1$, MedInsight-ViseurAI leads UW-Madison-AIR by 33.8 percentage points in Area, but this gap shrinks to only 7.2 points at $\delta_{10}$. This trend indicates that while all models capture the coarse vessel structure, significant differences emerge at finer structural scales, where precise boundary delineation becomes critical.

In the Classification Task, the top-performing team was UW-Madison-AIR with an average score of 62.3\%, while the second-best team, Haqan, achieved 54.3\%. The dilation rate analysis further reveals that UW-Madison-AIR achieves hepatic vessel Area scores ranging from 69.4\% ($\delta_1$) to 87.6\% ($\delta_{10}$), and portal vessel Area scores from 61.1\% to 76.6\% over the same range, confirming that hepatic vessels are consistently easier to segment than portal vessels. A comparison across metrics reveals that Area and Length scores are generally higher than IoU and NSD results for all teams. This discrepancy suggests that, while models struggle with exact voxel-wise overlap (IoU) and surface distances (NSD) for thin structures, they are more successful at capturing cross-sectional area (Area) and longitudinal continuity (Length). The increasing trend in the dilation rate further confirms that segmentation errors are often localized near vessel boundaries. As the tolerance region widens, these near-miss predictions are captured, yielding substantially higher scores.

To qualitatively compare performance, sample mask predictions, along with the ground truth and the baseline model results, are presented in Fig.~\ref{fig:Qualitative_results}. While the main vein structure was correctly identified across participants' models, smaller branches extending from the larger ones were poorly segmented. The predictions from UW-Madison-AIR showed higher alignment with the ground truth; however, there were still mispredictions in the non-liver regions for Patients 3, 4, 11, and 15 due to intensity similarities. This issue was also observed in other submissions. Nevertheless, our post-processing method for GLIMS effectively eliminated the non-liver predictions, resulting in fewer false-positive segmentations and achieving the highest overall scores, including 91.0\% in Area ($\delta_5$) and 89.3\% in Length ($\delta_5$). Over the full range of dilation rates, the GLIMS baseline achieves Area scores from 78.9\% ($\delta_1$) to 95.2\% ($\delta_{10}$) and Length scores from 80.1\% to 95.5\%, demonstrating consistently strong geometric performance across all tolerance levels. The quantitative results for the selected samples are shown in Fig.~\ref{fig:Quantitative_results}.


\section{Conclusions and Discussions}
\label{conc}
In this work, we introduced the VEELA dataset, a rigorously annotated resource designed to advance automated liver vessel segmentation in CTA images. The dataset aims to fill a critical gap in the field by providing a challenging, high-quality benchmark for assessing and developing vessel segmentation algorithms.

Our benchmarking experiments with Swin UNETR, GLIMS, and VSNet models provide important insights into the dataset's complexity and the capabilities of current state-of-the-art methods. Quantitative evaluations using clDice, IoU, NSD, Area, and Length metrics show that GLIMS consistently outperformed in both settings, with and without liver masks, achieving clDice scores of 76.9\% and 75.6\%, respectively, compared to Swin UNETR’s 69.6\% and 68.8\%. The inclusion of geometric metrics further highlights this performance, with GLIMS achieving 91.0\% in Area and 89.3\% in Length, demonstrating its superior ability to capture both the cross-sectional presence and longitudinal extent of the vessels.

Qualitative assessments further support that the GLIMS model produced precise, anatomically coherent segmentations, whereas Swin UNETR showed inconsistencies, particularly in predicting vessels outside the liver parenchyma. These discrepancies highlight models' sensitivity to anatomical context and suggest that integrating spatial priors, as in GLIMS, can significantly enhance performance.

Furthermore, we reported the team submissions to VEELA 2025 in Section \ref{challenge_results}. The results underscore both the promise and the challenge inherent in liver vessel segmentation. While advanced models demonstrate strong performance, accurately delineating thin, faint peripheral vessels remains difficult. Notably, the higher scores observed in Length compared to IoU across participants suggest that current models are more adept at following the vascular tree's path than precisely defining its volumetric boundaries.

Accurate evaluation is inherently challenging due to the intricate morphology of vascular structures and the necessity to preserve connectivity. In clinical applications, it is crucial that a segmentation method captures the entire network and accurately represents geometric properties such as thickness and extent. To this end, our framework integrates multiple complementary metrics, clDice, IoU, NSD, Area, and Length, to provide a holistic assessment. We are still searching for the best strategy for combining them to reflect the clinical experts' assessment.

Looking ahead, we anticipate that VEELA will serve as a valuable benchmark, catalyzing the development of more robust, context-aware methods. In future work, extending evaluations to include additional architectures, semi-supervised learning, and domain adaptation may provide further insights into overcoming the complex variability observed in clinical liver CTA imaging.

\section*{Declaration of Competing Interest}
The authors declare that they have no known competing financial interests or personal relationships that could have appeared to influence the work reported in this paper.

\section*{CRediT authorship contribution statement}
\textbf{Ziya Ata Yazıcı:} Methodology, Software, Formal analysis, Investigation, Writing - original draft, Writing - review \& editing, Visualization. 
\textbf{N. Sinem Gezer:} Data curation. 
\textbf{İlkay Öksüz:} Supervision, Writing - review \& editing. 
\textbf{İlker Özgür Koska:} Data curation.
\textbf{Tuğçe Toprak:} Investigation, Formal analysis, Software, Writing - original draft, Visualization
\textbf{Pervin Bulucu:} Writing - original draft. 
\textbf{Ufuk Beşenk:} Formal analysis, Writing - original draft, Writing - review \& editing. 
\textbf{A. Emre Kavur:} Conceptualization, Resources, Data curation.
\textbf{Pierre-Henri Conze:} Writing - original draft, Writing - review \& editing. 
\textbf{Hazım Kemal Ekenel:} Supervision, Writing - review \& editing.  
\textbf{Oğuz Dicle:} Data curation. 
\textbf{Mustafa Ege Şeker:} Methodology, Software. 
\textbf{Mustafa Said Kartal:} Methodology, Software. 
\textbf{Ariorad Moniri:} Methodology, Software. 
\textbf{Orhan Özkan:} Methodology, Software. 
\textbf{Osman Faruk Bayram:} Methodology, Software. 
\textbf{Hakan Polat:} Methodology, Software. 
\textbf{Musa Balcı:} Methodology, Software. 
\textbf{Ece Tuğba Cebeci:} Methodology, Software. 
\textbf{Baran Cılga:} Methodology, Software. 
\textbf{Kardelen Peçenek:} Methodology, Software. 
\textbf{M. Alper Selver:} Project administration, Conceptualization, Methodology, Formal analysis, Resources, Data curation, Writing - original draft, Writing - review \& editing.

\section*{Acknowledgments}
This work is supported by the Scientific and Technological Research Council of Türkiye (TUBITAK) under grant numbers 116E133, 123R036, and 123R064. Computing resources used in this work for baseline experiments were provided by the National Center for High Performance Computing of Türkiye (UHeM) under grant number 4021942025. The authors sincerely thank the IEEE Signal Processing Society for their financial and organizational support in providing the challenge prizes.


\bibliographystyle{model2-names.bst}\biboptions{authoryear}
\bibliography{refs}


\end{document}